\documentclass{article}
\usepackage{graphicx} 
\usepackage{tabularx}
\usepackage{amsmath}
\usepackage{xspace}
\usepackage{booktabs}
\usepackage{tcolorbox}

\usepackage{colortbl}
\usepackage{graphicx}
\usepackage{enumitem}
\usepackage{algorithm}
\usepackage{algpseudocode}

\usepackage{wrapfig}

\usepackage{amsmath}
\usepackage{amsfonts}
\usepackage{bm}
\usepackage{vcell}

\usepackage[utf8]{inputenc} 
\usepackage[T1]{fontenc}    
\usepackage{hyperref}       
\usepackage{url}            
\usepackage{booktabs}       
\usepackage{amsfonts}       
\usepackage{nicefrac}       
\usepackage{microtype}      
\usepackage{xcolor}         

\usepackage{booktabs}
\usepackage{graphicx}
\usepackage{subcaption}
\usepackage{tabularx}
\usepackage{graphicx}
\usepackage{amsmath}
\usepackage{amssymb}
\usepackage[preprint]{corl_2024} 

\usepackage{siunitx}
\DeclareSIUnit\px{px}
\DeclareSIUnit\fps{fps}
\DeclareSIUnit\dof{DoF}

\title{\raisebox{-0.1\height}{\includegraphics[width=0.65cm]{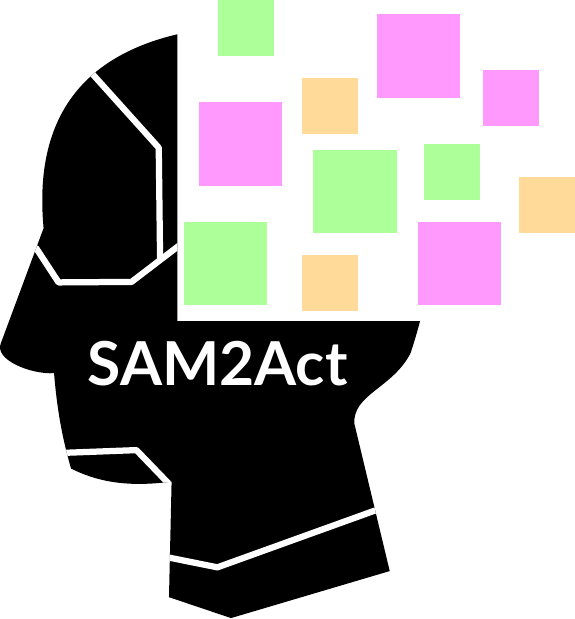}}SAM2Act: Integrating Visual Foundation Model with A Memory Architecture for Robotic Manipulation}

\author{ Haoquan Fang~$^{1}$ \hspace{4px} Markus Grotz~$^{1}$ \hspace{4px} Wilbert Pumacay~$^2$ \hspace{4px} Yi Ru Wang~$^1$ \\
\textbf{Dieter Fox~$^{1,3*}$} \hspace{4px} \textbf{Ranjay Krishna~$^{1,4*}$}\hspace{4px} \textbf{Jiafei Duan~$^{1,4}$\thanks{Equal advising}}
\\
$^1$University of Washington \hspace{6px} $^2$Universidad Católica San Pablo \hspace{6px} \\
$^3$NVIDIA $^4$Allen Institute for Artificial Intelligence \hspace{6px}\\
[1em]
\large\textbf{\href{https://sam2act.github.io/}{sam2act.github.io}
}
}

\begin{document}
\maketitle


\begin{abstract}
Robotic manipulation systems operating in diverse, dynamic environments must exhibit three critical abilities: multitask interaction, generalization to unseen scenarios, and spatial memory. While significant progress has been made in robotic manipulation, existing approaches often fall short in generalization to complex environmental variations and addressing memory-dependent tasks. To bridge this gap, we introduce \textbf{SAM2Act}, a multi-view robotic transformer-based policy that leverages multi-resolution upsampling with visual representations from large-scale foundation model. SAM2Act achieves a state-of-the-art average success rate of \textbf{86.8\% across 18 tasks} in the RLBench benchmark, and demonstrates robust generalization on \texttt{The Colosseum} benchmark, with only a \textbf{4.3\% performance gap} under diverse environmental perturbations. Building on this foundation, we propose \textbf{SAM2Act+}, a memory-based architecture inspired by SAM2, which incorporates a memory bank, an encoder, and an attention mechanism to enhance spatial memory. To address the need for evaluating memory-dependent tasks, we introduce \textbf{\texttt{MemoryBench}}, a novel benchmark designed to assess spatial memory and action recall in robotic manipulation. SAM2Act+ achieves an average success rate of \textbf{94.3\% on memory-based tasks} in \texttt{MemoryBench}, significantly outperforming existing approaches and pushing the boundaries of memory-based robotic systems.
\end{abstract}
\vspace{-1em}

\keywords{Robotics Manipulation, Multiview Robotics Transformer, Imitation Learning, Memory-based Architecture, Behavior Cloning, Generalization} 

\section{Introduction}
\label{sec:intro}

\begin{figure*}[!ht]
    \vskip 0.2in
    \begin{center}
        \centerline{\includegraphics[width=\textwidth]{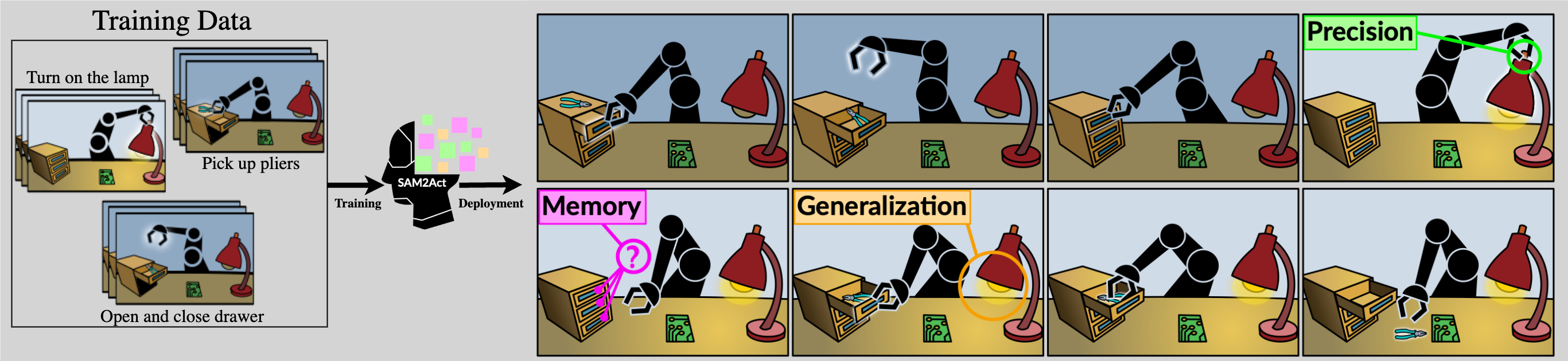}}
        \caption{SAM2Act is a multi-view, language-conditioned behavior cloning policy trained with fewer demonstrations. Given a language instruction, it can execute high-precision tasks, such as turning the tiny knob on the lamp. It also generalizes to various environmental variations, such as changes in lighting conditions. Through further training with our proposed memory architecture, it now evolves into SAM2Act+, which is now capable of solving tasks that require implicit spatial memory—such as remembering where the robot previously stored the pliers, as depicted in the above figure.}
        \label{fig:teaser}
    \end{center}
    \vskip -0.2in
\end{figure*}

The world in which we live is diverse and constantly changing, encompassing a wide variety of objects, scenes, and environmental conditions. Consider the seemingly simple task of following a recipe when cooking: we can seamlessly perform the action of picking it up and sprinkling it into the pan, recognize salt even if it comes in different types of container, and remember whether we have already added salt. Humans excel in such environments because they can interact with their surroundings to achieve specific goals, generalize to unseen scenarios, and retain knowledge from past experiences \cite{smith2005development}. These abilities—multitask interaction, generalization, and memory—serve as guiding principles for developing robotic systems capable of operating in similarly complex environments.

Significant progress has been made in robotic manipulation through prior work. Early methods, such as the Transporter Network \cite{zeng2021transporter} and CLIPort \cite{shridhar2022cliport}, demonstrated effective 2D action-centric manipulation but were limited in their ability to handle spatially complex tasks. More recent approaches, such as PerAct \cite{shridhar2023perceiver} and RVT \cite{goyal2023rvt}, have pushed toward 3D-based manipulation. PerAct employs a multitask transformer that interprets language commands and predicts keyframe poses, achieving strong results across a variety of tasks. RVT builds on this foundation by adopting a 2.5D representation, improving training efficiency and inference speed. Its successor, RVT-2, further enhances performance with a coarse-to-fine strategy, increasing precision for high-accuracy tasks. Despite these advances, important challenges remain, including improving multitask performance, enhancing generalization to novel environment configurations, and integrating memory mechanisms for tasks requiring episodic recall.

We introduce SAM2Act, a multi-view robotics transformer-based policy that enhances feature representation by integrating multi-resolution upsampling with visual embeddings from large-scale foundation models. Built on the RVT-2 multi-view transformer, SAM2Act achieves strong multitask success and generalization. Building on this foundation, we introduce SAM2Act+, which incorporates a memory-based architecture inspired by SAM2's approach. Using a memory bank, an encoder, and an attention mechanism, SAM2Act+ enables episodic recall to solve spatial memory-dependent manipulation tasks. We evaluate SAM2Act and SAM2Act+ using \texttt{MemoryBench}, a new benchmark suite that tests policies' spatial memory capabilities and the ability to retain and recall past actions. SAM2Act+ achieves an average success rate of 94.3\% across all tasks on \texttt{MemoryBench}, with an average accuracy of 94.3\%, outperforming next highest baseline by a huge margin of 39.3\%. Furthermore, we assess the generalization capabilities of SAM2Act on \texttt{The Colosseum} \cite{pumacay2024colosseum}, a benchmark designed to test robotic manipulation under various environmental perturbations. SAM2Act demonstrates robust performance on \texttt{The Colosseum} with an average decrease of 4.3\% across all perturbations, highlighting its ability to generalize effectively in diverse and challenging scenarios. Lastly, our approach outperforms the baseline methods in real-world evaluations while exhibiting comparable generalization and spatial memory capabilities. 

In summary, this work makes three key contributions. First, we introduce a \textbf{novel model formulation} that leverages visual foundation models to solve \textbf{high-precision, memory-dependent manipulation tasks}. Second, we propose \texttt{MemoryBench}, a evaluation benchmark for \textbf{assessing spatial memory in behavior cloning models}. Finally, we present \textbf{empirical results and insights} on the model’s performance across both simulation and real-world tasks.
\section{Related Work}
\label{sec:related_work}

\subsection{3D-based Robotic Transformer for Manipulation}

2D-based methods \cite{zhao2023learning,chi2023diffusion,zeng2021transporter,brohan2022rt,shridhar2022cliport} are effective for simple pick-and-place tasks due to fast training, low hardware requirements, and minimal computational cost. However, they depend on pretrained image encoders and fail in tasks requiring high precision, robust spatial interaction, or resilience to environmental and camera variations \cite{pumacay2024colosseum}. Recent work addresses these limitations with 3D perception. Methods like PolarNet \cite{chen2023polarnet}, M2T2 \cite{yuan2023m2t2}, and Manipulate-Anything \cite{duan2024manipulate} reconstruct point clouds, while C2F-ARM \cite{james2022coarse} and PerAct \cite{shridhar2023perceiver} use voxel-based 3D representations. Act3D \cite{gervet2023act3d} and ChainedDiffuser \cite{xian2023chaineddiffuser} adopt multi-scale 3D features. RVT \cite{goyal2023rvt} introduces 2.5D multi-view images for faster training, refined by RVT-2 \cite{goyal2024rvt} with a coarse-to-fine architecture for improved precision. Our work, SAM2Act, combines RVT-2’s spatial reasoning with enhanced virtual images from the SAM2 visual encoder, achieving high precision and generalization across diverse tasks.

\subsection{Visual Representations for Robot Learning}

Robotics research heavily relies on visual representations from computer vision to process high-dimensional inputs and improve policy learning. Visual representations are integrated into robot learning through pre-training \cite{majumdar2023we,ma2022vip,nair2022r3m}, co-training \cite{laskin2020reinforcement,yarats2021image,laskin2020curl,shang2024theia}, or frozen encoders \cite{shah2021rrl,wang2022vrl3,zhang2024sam}, all of which effectively support policy training. These representations also enhance invariance, equivariance, and out-of-distribution generalization \cite{wang2022equivariant,pumacay2024colosseum,dasari2023unbiased}. Notably, object-centric visual representations (e.g. from SAM) are shown to be even more useful and relevant for robotic manipulation and control \cite{shi2024pocr, qian2024hodor}. Specifically, SAM-E \cite{zhang2024sam} demonstrates the use of a pre-trained SAM encoder for robotic manipulation by leveraging image embeddings for policy learning. Expanding on this, our approach employs the SAM2 visual encoder to generate image embeddings for robotic transformers and utilizes its multi-resolution features to improve convex upsampling for next-action prediction.

\subsection{Memory in Robotics}

Memory is a fundamental component of human cognition, and equipping generalist robotic agents with episodic and semantic memory is crucial for enabling them to perform complex tasks effectively \cite{jockel2008towards}. Early research on memory in robotics primarily addressed navigation tasks, relying on semantic maps that were often constrained in scope \cite{henry2012rgb,bowman2017probabilistic,chaplot2020object}. Other work explicitly model the memory and its representation for a robot cognitive architecture \cite{peller2023memory}. Recent advancements leverage representations derived from vision-language models (VLMs) and Large Vision Models (LVMs), utilizing voxel maps or neural feature fields to encode, store, and retrieve information \cite{huang2024copa,huang2023voxposer,duan2024manipulate,liu2024dynamem}. Alternative methods represent semantic memory for manipulation tasks using Gaussian splats to encode spatial information \cite{kerbl20233d,shorinwa2024splat}. Recent work \cite{huang2024out} employs transformer-based relational memory on partial-view point clouds—augmented with object discovery and tracking—to robustly handle occlusions, novel and reappearing objects, and diverse distractors, outperforming implicit-memory baselines in both simulation and real-world experiments. In contrast, our approach draws inspiration from the framework of Partially Observable Markov Decision Processes (POMDPs) \cite{lauri2022partially}, incorporating memory directly into the training process. By integrating spatial memory from past actions into the agent’s belief state, we enhance the robustness and adaptability of learned policies.
\section{\texttt{MemoryBench}: A Memory Benchmark for Robotic Manipulation}
\label{sec:memorybench}

\begin{figure*}[!ht]
    \vskip 0.2in
    \begin{center}
        \centerline{\includegraphics[width=\textwidth]{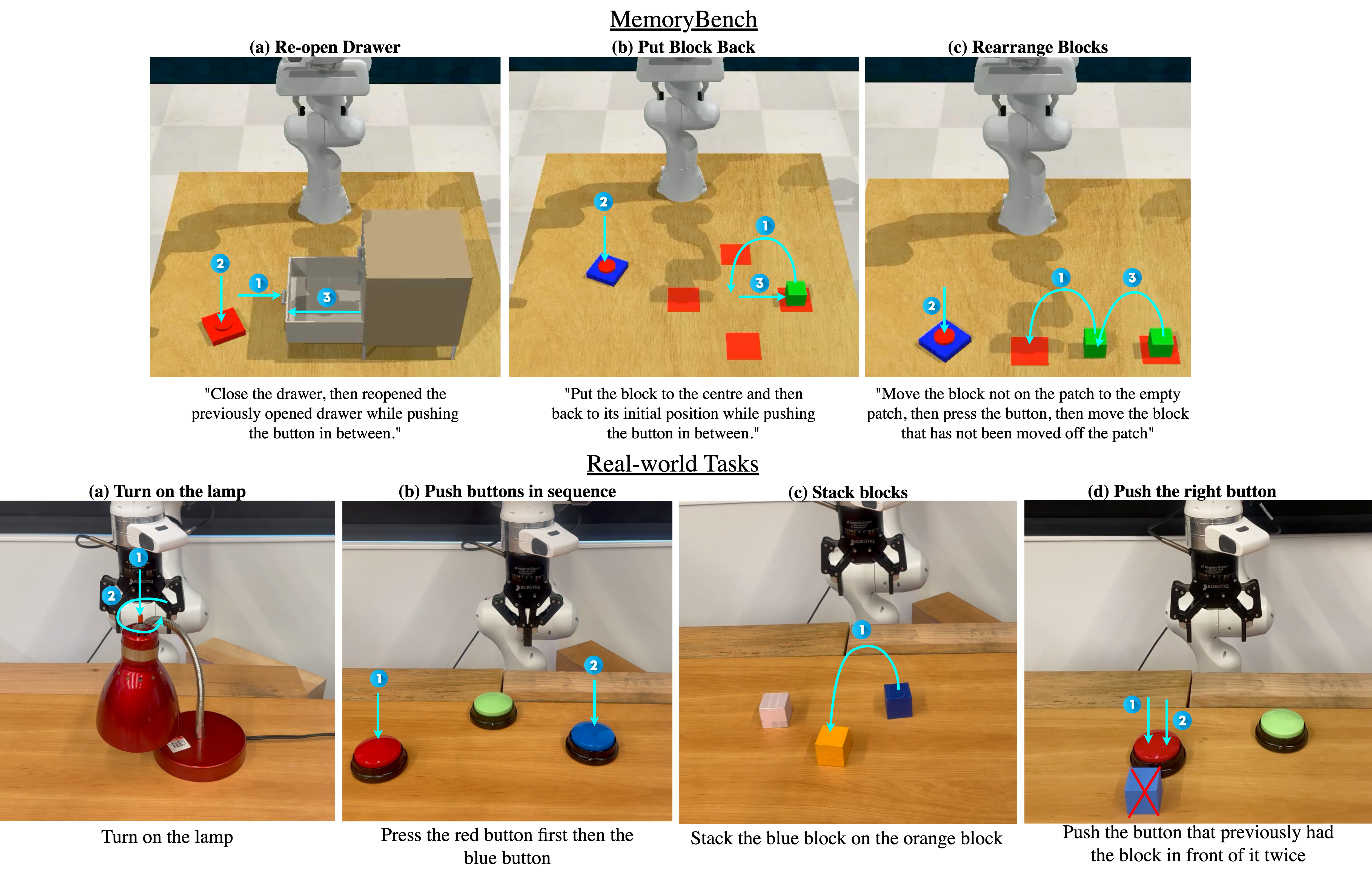}}
        \caption{\textbf{Simulation and Real Tasks.} We demonstrate the effectiveness of SAM2Act+ in solving memory-based tasks by evaluating it against baselines on the three benchmark memory tasks (shown at the top). Additionally, we validate our approach using a Franka Panda robot on four real-world tasks (shown at the bottom), including tests under out-of-distribution perturbations.}
        \label{fig:tasks}
    \end{center}
    \vskip -0.2in
\end{figure*}

We introduce \texttt{MemoryBench}, a benchmark designed to systematically evaluate the spatial memory capabilities of robotic manipulation policies. In \autoref{sec:memorybench:design}, we begin by outlining the logic and rules behind task design. We will then describe the tasks we have developed in \autoref{sec:memorybench:tasks}.

\subsection{Task Design}
\label{sec:memorybench:design}

Unlike standard RLBench tasks \cite{james2020rlbench}, many of which involve long-horizon scenarios, our tasks are specifically designed to require spatial memory. Without such memory, the agent would be forced to rely on random actions. To create these tasks, we intentionally violate the Markov assumption, which states that in a Markov Decision Process (MDP), the next observation depends solely on the current observation and action:
\[
    P\bigl(o_{t+1} \mid o_1, a_1, \dots, o_t, a_t\bigr)
    \;=\;
    P\bigl(o_{t+1} \mid o_t, a_t\bigr).
\]
This assumption implies that knowing only \( o_t \) and \( a_t \) is sufficient to predict \( o_{t+1} \). However, in our tasks, we design scenarios where two distinct action histories lead to the same observation \( o_t \), but require different subsequent actions. This forces the agent to recall which action history led to \( o_t \) to perform the correct next action. Furthermore, we standardized the language instructions to prevent unintentional leakage of spatial information that could aid the model in memory-based tasks. These principles guided the development of our spatial memory-based tasks.

\subsection{Spatial Memory-based Tasks}
\label{sec:memorybench:tasks}

\texttt{MemoryBench} extends the RLBench simulator to provide scripted demonstrations for three spatial memory tasks: \texttt{reopen\_drawer}, \texttt{put\_block\_back}, and \texttt{rearrange\_block}. Each task is designed to evaluate a specific aspect of spatial memory and adheres to the principles outlined in Section~\ref{sec:memorybench:design}. To introduce complexity, these tasks include two to four variations and additional steps—such as pressing a button mid-sequence—that disrupt the Markov property. This forces the agent to rely on memory rather than solely on immediate observations.

The \texttt{reopen\_drawer} task evaluates the agent’s ability to recall 3D spatial information along the z-axis. Initially, one of three drawers (top, middle, or bottom) is open. The agent must close the open drawer, press a button on the table, and then reopen the same drawer. After the button is pressed, all drawers are closed, and the scene becomes visually indistinguishable, requiring the agent to use memory to identify the correct drawer. This task tests the agent’s ability to recall spatial states over a temporal sequence. The \texttt{put\_block\_back} task tests the agent’s ability to remember 2D spatial information on the x-y plane. Four red patches are placed on a table, with a block initially positioned on one of them. The agent should move the block to the center of the patches, press a button, and return the block to its original position. The agent must rely on its memory of the block’s initial location to succeed, demonstrating its capability to encode and retrieve 2D spatial information.

The \texttt{rearrange\_block} task evaluates the agent’s ability to perform backward reasoning by recalling and reversing prior actions. Initially, one block is placed on one of two red patches, while the other patch remains empty. A second block is positioned at the center of both patches. The agent must move the second block to the empty patch, press a button, and then relocate the first block off its patch. Successfully completing this task requires the agent to determine which block to move without having interacted with the correct one in previous actions, thereby testing its capacity for backward spatial memory reasoning. These tasks collectively evaluate both forward and backward spatial reasoning across 3D (z-axis) and 2D (x-y plane) spaces. By introducing non-Markovian elements, they emphasize the need for memory representations to solve complex sequential decision-making problems (more details in \autoref{app:memorybench}).

\section{Method}
\label{sec:method}

\begin{figure*}[!t]
    \vskip 0.2in
    \begin{center}
        \centerline{\includegraphics[width=\textwidth]{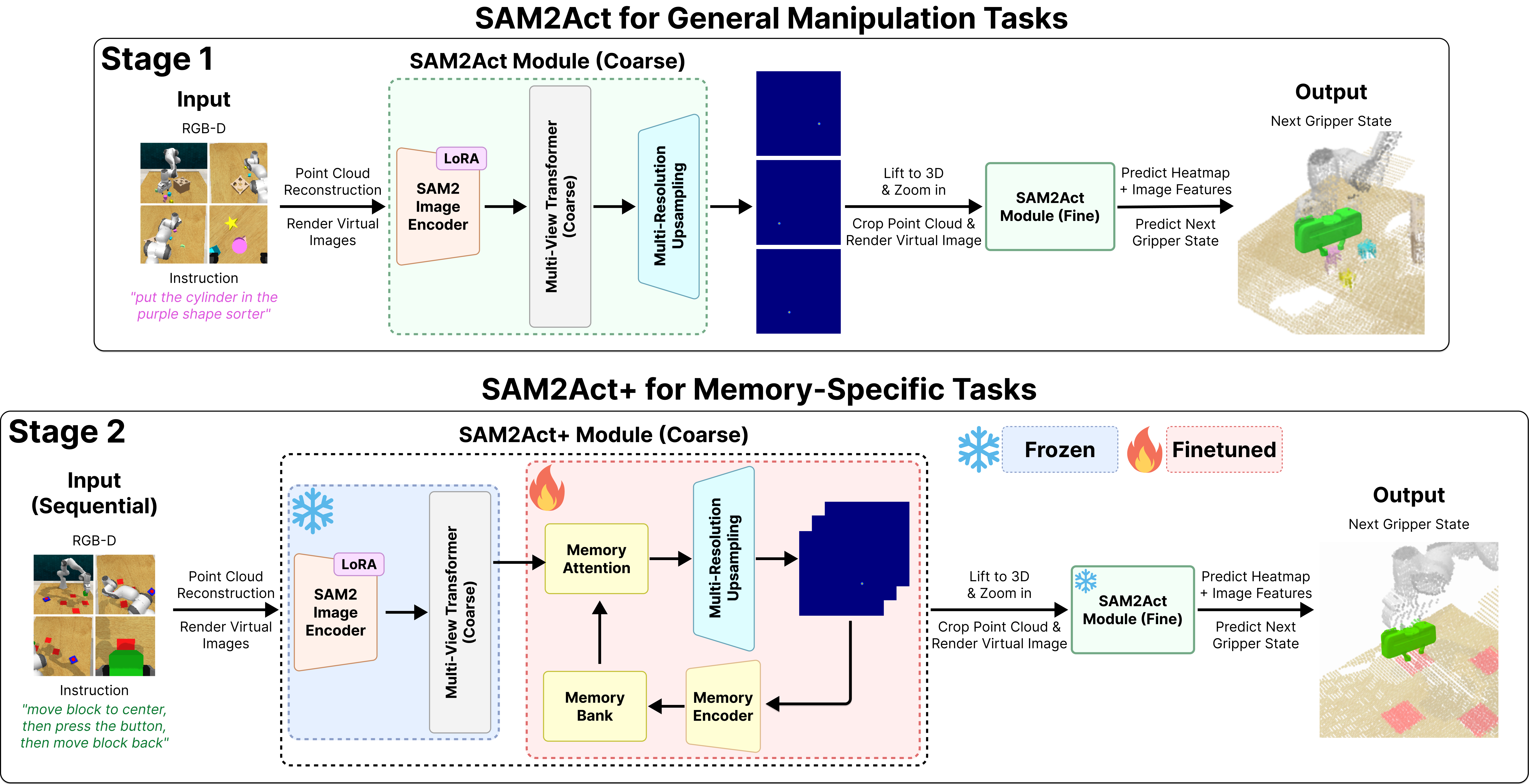}}
        \caption{\textbf{Overview of the SAM2Act (top) and SAM2Act+ (bottom) architectures.} The SAM2Act architecture leverages the SAM2 image encoder to generate prompt-conditioned, multi-resolution embeddings, fine-tuned with LoRA for efficient adaptation to manipulation tasks. A multi-view transformer aligns spatial coordinates with language instructions, while a cascaded multi-resolution upsampling mechanism refines feature maps and generates accurate translation heatmaps. SAM2Act+ extends this architecture by incorporating memory-based components, including the Memory Encoder, Memory Attention, and Memory Bank, into the coarse branch. These components enable memory-driven reasoning by processing historical heatmaps and integrating prior observations, allowing the agent to predict actions based on stored contextual information. Observations are reconstructed into point clouds, rendered into three virtual images, and lifted into 3D translation points, enabling precise spatial reasoning across both architectures.}
        \label{fig:sam2act}
    \end{center}
    \vskip -0.2in
\end{figure*}

Our method, SAM2Act, enables precise 3D manipulation with strong generalization across environmental and object-level variations. Building upon the RVT-2 framework \cite{goyal2024rvt}, SAM2Act introduces key architectural innovations that enhance visual feature representation and task-specific reasoning. The architecture reconstructs a point cloud of the scene, renders it from virtual cameras at orthogonal views, and employs a two-stage multi-view transformer (coarse-to-fine) to predict action heatmaps. The coarse branch generates zoom-in heatmaps to localize regions of interest, while the fine branch refines these into precise action heatmaps. SAM2Act leverages the pre-trained SAM2 encoder \cite{ravi2024sam} to extract multi-resolution image embeddings, which are further refined through the multi-resolution upsampling technique to predict accurate translation heatmaps with minimal information loss. To address tasks requiring spatial memory, SAM2Act+ extends the SAM2Act architecture by incorporating memory-based components. These include Memory Bank, Memory Encoder, and Memory Attention, enabling the model to encode historical actions and condition current observations. This memory-based policy enhances the agent’s ability to predict actions based on past contextual information, significantly improving performance in tasks that require sequential decision-making. 

In the following sections, we detail the SAM2Act architecture (\autoref{sec:method:sam2act}), including its multi-resolution upsampling mechanism (\autoref{fig:upsample}). We also present the SAM2Act+ extension, which integrates memory-based components for solving spatial memory tasks (\autoref{sec:method:sam2act+}).

\subsection{SAM2Act: Multi-Resolution Upsampling for Enhanced Visual Feature Representation}
\label{sec:method:sam2act}

\begin{figure}[!t]
    \vskip 0.2in
    \begin{center}
        \centerline{\includegraphics[width=\linewidth]{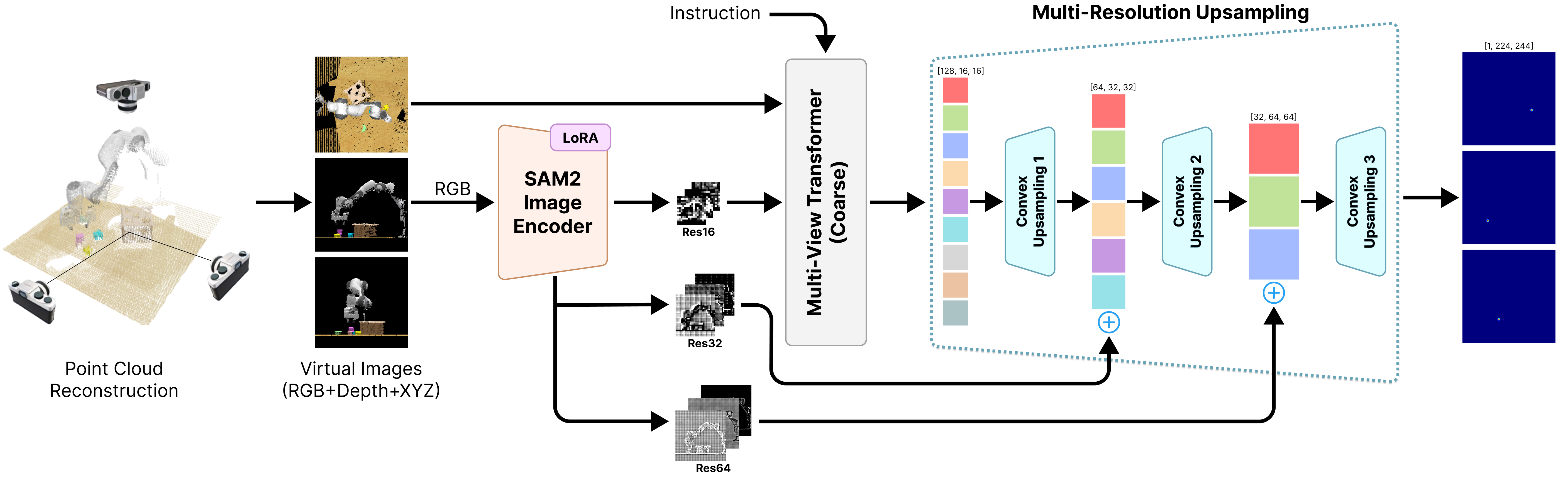}}
        \caption{\textbf{SAM2Act Module and multi-resolution upsampling mechanism.} A cascade of three convex upsamplers processes feature maps at increasing resolutions, integrating multi-resolution embeddings from the SAM2 image encoder through elementwise addition and layer normalization. The upsamplers progressively refine features, doubling spatial dimensions at each stage, to generate accurate translation heatmaps while capturing fine-grained spatial details critical for manipulation tasks.}
        \label{fig:upsample}
    \end{center}
    \vskip -0.2in
\end{figure}

A distinctive feature of SAM2Act is the incorporation of the SAM2Act Module into the manipulation backbone for training, as illustrated in \autoref{fig:upsample}. The coarse and fine SAM2Act Modules share the same architecture, with the fine branch generating additional features to predict actions beyond translation, while the coarse branch focuses exclusively on translation. Point-cloud representations are reconstructed from raw image inputs, and virtual images are generated from three viewpoints using virtual cameras. Instead of directly inputting these images into the multi-view transformer, their RGB channels are duplicated and processed by the SAM2 \cite{ravi2024sam} image encoder, which produces object-centric multi-resolution embeddings. These embeddings, generated at three resolution levels, are combined with virtual images containing RGB, depth, 3D translation coordinates, and language instructions before being fed into the multi-view transformer. Details of how we adapt the MVT can be found in \autoref{app:architecture}.

To adapt the SAM2 image encoder to our domain, we fine-tune it using Low-Rank Adaptation (LoRA) \cite{hu2021lora} with a default rank of 16, which enables domain adaptation with minimal computational cost while maintaining model efficiency. Additionally, to fully leverage the multi-resolution embeddings produced by the SAM2 image encoder, we introduce a multi-resolution upsampling method. This method uses the embeddings as auxiliary inputs to enhance the generation of translation heatmaps, thereby improving spatial precision and overall system performance. The multi-resolution upsampling mechanism, also detailed in \autoref{fig:upsample}, leverages cascaded convex upsamplers to progressively refine feature maps across resolutions. Let \(X^l \in \mathbb{R}^{B \times C^l \times H^l \times W^l}\) denote the feature maps at stage \(l\) and \(E^l \in \mathbb{R}^{B \times C^l \times H^l \times W^l}\) the corresponding multi-resolution embedding from SAM2. Also let \(U(\cdot)\) denote the upsampling operator that doubles the spatial dimensions. The feature maps are updated at each stage as follows:
\[
X^{l+1} = \mathrm{LayerNorm}\bigl(U(X^l) \,\oplus\, E^l \bigr),
\]
where \(\oplus\) represents element-wise addition. The upsampling operator \(U\) is defined as:
\[
U: \mathbb{R}^{B \times C^l \times H^l \times W^l}
\,\rightarrow\,
\mathbb{R}^{B \times (C^l/2) \times (2H^l) \times (2W^l)}.
\]
At each stage, the output of the upsampler is combined with the corresponding multi-resolution embedding \(E^l\) from the SAM2 encoder, ensuring alignment between the multi-resolution features and the decoder’s spatial refinement process. A layer normalization step follows each addition to stabilize training and maintain feature coherence. This results in direct integration of the embeddings into the translation heatmap generation process. The cascading structure refines features across multiple resolutions, capturing fine-grained spatial details critical for manipulation tasks.

\subsection{SAM2Act+: Action Memory Architecture for Improved Spatial Awareness in Past Observations}
\label{sec:method:sam2act+}

\begin{algorithm}[tb]
   \caption{Forward Pass of SAM2Act+ Module}
   \label{algo:forward}
   \begin{algorithmic}[1] 
       \State \textbf{Initialize:} Number of steps $N$, maximum number of memories $M$, number of views $V$, empty memory bank $Q$ with $V$ separate FIFO queues, input $X$
       \For{$i = 1$ \textbf{to} $N$}
           \For{$j = 1$ \textbf{to} $V$}
               \State Get embeddings $\mathcal{E}_{raw}$ from MVT $T_{mv}(X_{j})$
               \State Retrieve past memories $\mathcal{M}_{old}$ from $Q[j]$
               \State Get memory-conditioned embeddings $\mathcal{E}_{mem}$ from Memory Attention $T_{mem}(\mathcal{E}_{raw}, \mathcal{M}_{old})$
               \State Predict translation heatmap $\mathcal{H}$ with upsampler $U(\mathcal{E}_{mem})$
               \State Encode new memory $\mathcal{M}_{new}$ using Memory Encoder $E_{mem}(\mathcal{H}, \mathcal{E}_{raw})$
               \State Store new memory $Q[j] \gets Q[j] \cup \{\mathcal{M}_{new}\}$
               \If{$|Q[j]| = M$}
                   \State $Q[j] \gets Q[j]_{2:n}$
               \EndIf
           \EndFor
       \EndFor
   \end{algorithmic}
\end{algorithm}

To extend the SAM2Act architecture (\autoref{sec:method:sam2act}) with memory-based capabilities inspired by SAM2, we introduce SAM2Act+, a task-specific variant designed for solving memory-based tasks. SAM2Act+ integrates the three core memory components from SAM2—\textit{Memory Attention, Memory Encoder, and Memory Bank}—into the coarse branch of SAM2Act. Originally developed for object tracking in SAM2, these components are adapted to align with the needs of SAM2Act+, enabling the agent to retain prior actions and observations for sequential decision-making. In SAM2, the Memory Encoder processes predicted object masks, while the Memory Attention module fuses image embeddings with positional information from previous frames. SAM2Act+ adopts a similar structure: the predicted heatmaps, which serve as binary indicators of spatial positions in the image, function analogously to object masks. This conceptual alignment ensures a seamless integration of memory mechanisms, allowing the agent to leverage stored information to predict subsequent actions based on historical context. A detailed description of the Memory Attention and Memory Encoder modules can be found in \autoref{app:architecture}.

\textbf{Architecture.} The SAM2Act+ architecture is illustrated in \autoref{fig:sam2act}. After pretraining SAM2Act in Stage 1, we freeze the SAM2 image encoder and the multi-view transformer in the coarse branch, as these components effectively generate robust embeddings for multi-view images in manipulation tasks. We also freeze the entire fine branch, given its proven ability to predict fine-grained actions accurately. The reason why we only fine-tune the coarse branch is because it focuses on generating heatmaps that provide richer contextual information for recalling past actions. The fine branch, in contrast, primarily emphasizes small objects or localized regions, which typically contain less information relevant to memory-based tasks.

\textbf{Training.} To train SAM2Act+, we fine-tune the coarse branch by integrating the three memory components (and train them from scratch) with the multi-resolution upsampling module. During fine-tuning, consecutive action keyframes are sampled as input, training the multi-resolution upsampler to predict new translations conditioned on memory. The memory components function similarly to their implementation in SAM2 for object tracking, with one key distinction: the input to the Memory Encoder. Instead of using image embeddings from the SAM2 image encoder, we input feature embeddings generated by the multi-view transformer (not conditioned by memory). This adaptation ensures that memory encoding incorporates multi-view information while maintaining independence in handling stored representations. Virtual images are treated independently during memory encoding and attention, with each view’s memory encoded separately. Feature embeddings from each view are attended to using their corresponding stored memories, preserving spatial and contextual alignment while leveraging fused multi-view information. This structured approach prevents cross-view interference and enhances the model’s ability to reason over sequential tasks. The memory-based forward pass for SAM2Act+ is outlined in \autoref{algo:forward}. By incorporating the memory mechanism, SAM2Act+ enhances performance in scenarios requiring long-term reasoning, enabling the agent to make informed decisions based on historical context.

\section{Experiments}
\label{sec:experiments}

We study SAM2Act and SAM2Act+ in both simulated and real-world environments. Specifically, we are interested
in answering the following questions: 
\vspace{-0.3cm}
\begin{itemize}
    \setlength\itemsep{-0.2em}
    \item [\S~\ref{sec:experiments:baselines}] How does SAM2Act compare with state-of-the-art 3D manipulation policies?
    \item [\S~\ref{sec:experiments:generalization}] Can SAM2Act generalize across object and environmental perturbations?
    \item [\S~\ref{sec:experiments:memory}] Can SAM2Act+ solve spatial memory-based tasks that other baselines cannot?
    \item [\S~\ref{sec:experiments:real}] How well does SAM2Act and SAM2Act+ perform on real-world tasks?
\end{itemize}

\subsection{Experimental Setup}
\label{sec:experiments:setup}
We benchmark SAM2Act in both simulated and real-world environments. The simulated environments serve as a controlled platform to ensure reproducible and fair comparisons. The real-world experiments demonstrate the applicability of the method to real-world settings. Section \ref{sec:experiments:setup} details our experimental setup and outlines the evaluation methodology. Training details can be found in \autoref{app:training}.

\textbf{Simulation Setup.} All simulated experiments were conducted in the CoppeliaSim environment via PyRep, using a 7-DoF Franka Emika Panda robot in a tabletop setting. Observations were captured from five RGB-D cameras—front, left shoulder, right shoulder, overhead and wrist—each at $\SI{128}{\px} \times \SI{128}{\px}$. The robot receives a keyframe specifying translation and quaternion orientation and utilizes an OMPL-based motion planner to move to the target pose.

\textbf{Real-robot Setup.} We validate SAM2Act in real-world scenarios using a Franka Emika Panda robot with a Robotiq 2F-85 gripper and a exocentric Intel RealSense D455 depth sensor (more in \autoref{app:real_world}). We study four manipulation tasks, aligning three with RVT-2 for comparison and introducing a new memory-based task. We use the software stack as in \cite{grotz2024peract}. For each task, we collect 10–15 demonstrations via kinesthetic teaching and scripted execution with scene and object variations. As in \autoref{fig:tasks}, we evaluate SAM2Act against RVT-2 for tasks (a)–(c) and SAM2Act+ for memory task (d). Each task undergoes 10 in-distribution and 10 out-of-distribution trials, including environmental perturbations, measuring total success.

\textbf{18 RLBench \& \texttt{MemoryBench} Tasks.} To evaluate the general performance of SAM2Act and the memory capabilities of SAM2Act+, we conducted simulation experiments on two benchmarks: a subset of 18 tasks from RLBench and \texttt{MemoryBench}. RLBench is a standard multi-task manipulation benchmark, from which we selected 18 tasks well-studied in prior work. \texttt{MemoryBench} is a curated set of three tabletop manipulation tasks in CoppeliaSim that require the trained policy to have both semantic and spatial memory of past scenes and actions. In both benchmarks, each task is defined by a language instruction with 2–60 variations (e.g., handling objects, locations, and colors). We collected 100 demonstrations per task for training and held out 25 unseen demonstrations per task for testing. All policies are evaluated four times to obtain standard deviations. Tasks details can be found in \autoref{app:rlbench} and \autoref{app:memorybench}.

\textbf{3D Baselines.} We benchmark SAM2Act and SAM2Act+ against the current state-of-the-art 3D next-best-pose prediction model, RVT-2. RVT-2 is a multi-view robotics transformer that leverages a coarse-to-fine approach on the constructed point cloud to predict the next best action heatmap. We also compare with RVT \cite{goyal2023rvt}, PerAct \cite{shridhar2023perceiver}, and SAM-E \cite{zhang2024sam}.

\subsection{Performances Across 18 RLBench Tasks}
\label{sec:experiments:baselines}
\autoref{tab:rlbench} compares SAM2Act with prior keyframe-based 3D BC methods on the RLBench benchmark.  
Overall, SAM2Act achieves an average success rate of \textbf{86.8\%}$\pm$0.5, surpassing the previous best (RVT-2) by \textbf{5.4\%}. A closer look at individual tasks reveals that SAM2Act ranks \textbf{first in 9 out of 18 tasks} and remains highly competitive in \textbf{7 others}, coming within \textbf{one successful attempt or 4\%} of the best performance. These tasks include Close Jar, Drag Stick, Meat Off Grill, Place Wine, Screw Bulb, Sweep to Dustpan, and Turn Tap.  The largest margin of improvement occurs in \texttt{Insert Peg}, where SAM2Act \textbf{exceeds RVT-2 by 44\% (approximately 2.1$\times$)}, and in \texttt{Sort Shape}, where it outperforms RVT-2 by 29\%. Both tasks require precise manipulation, underscoring the effectiveness of SAM2Act’s multi-resolution upsampling strategy.  These results establish SAM2Act as a \textbf{leading policy for complex 3D tasks}, highlighting its ability to handle high-precision manipulations - an area where prior methods have struggled. Ablation studies are performed on SAM2Act in \autoref{app:sam2act_ablation}.

\begin{table*}[h!]
\centering
\caption{\textbf{Multi-Task Performance on RLBench.} We evaluate 18 RLBench tasks \cite{james2020rlbench}, reporting success rates across all tasks among 3D keyframe-based behavior cloning (BC) policies. We report stats of 4 evaluations for SAM2Act. Our method, SAM2Act, outperforms all baselines, achieving a significant performance margin of 5.8\% over RVT-2 \cite{goyal2024rvt}, the prior state-of-the-art 3D keyframe-based BC policy. Against all existing approaches, SAM2Act remains the state-of-the-art. See full comparisons in \autoref{app:rlbench_full}.}
\resizebox{\textwidth}{!}{%
\begin{tabular}{lccccccccccccc}

\toprule
\textbf{Method}     & \textbf{Avg. Success $\uparrow$} & \textbf{Avg. Rank $\downarrow$} & \textbf{Close Jar} & \textbf{Drag Stick} & \textbf{Insert Peg} & \textbf{Meat off Grill} & \textbf{Open Drawer} & \textbf{Place Cups} & \textbf{Place Wine} & \textbf{Push Buttons} \\
\midrule
PerAct \cite{shridhar2023perceiver}             & 49.4 $\pm$ 4.3                   & 4.6                               & 55.2 $\pm$ 4.7     & 89.6 $\pm$ 4.1      & 5.6 $\pm$ 4.1       & 70.4 $\pm$ 2.0          & 88.0 $\pm$ 5.7       & 2.4 $\pm$ 3.2       & 44.8 $\pm$ 7.8      & 92.8 $\pm$ 3.0        \\
RVT \cite{goyal2023rvt}                & 62.9 $\pm$ 3.7                   & 3.6                               & 52.0 $\pm$ 2.5     & 99.2 $\pm$ 1.6      & 11.2 $\pm$ 3.0      & 88.0 $\pm$ 2.5          & 71.2 $\pm$ 6.9       & 4.0 $\pm$ 2.5       & 91.0 $\pm$ 5.2      & \textbf{100.0} $\pm$ 0.0       \\
RVT-2 \cite{goyal2024rvt}                & 81.4 $\pm$ 3.1                   & 1.9                               & \textbf{100.0} $\pm$ 0.0    & 99.0 $\pm$ 1.7      & 40.0 $\pm$ 0.0      & \textbf{99.0} $\pm$ 1.7          & 74.0 $\pm$ 11.8      & 38.0 $\pm$ 4.5      & \textbf{95.0} $\pm$ 3.3      & \textbf{100.0} $\pm$ 0.0       \\
SAM-E \cite{zhang2024sam}               & 70.6 $\pm$ 0.7                   & 2.6                               & 82.4 $\pm$ 3.6     & \textbf{100.0} $\pm$ 0.0     & 18.4 $\pm$ 4.6      &  95.2 $\pm$ 3.3         & \textbf{95.2} $\pm$ 5.2       & 0.0 $\pm$ 0.0       & 94.4 $\pm$ 4.6      & \textbf{100.0} $\pm$ 0.0       \\
SAM2Act (Ours)      & \textbf{86.8} $\pm$ 0.5                   & \textbf{1.8}                               & 99.0 $\pm$ 2.0     & 99.0 $\pm$ 2.0      & \textbf{84.0} $\pm$ 5.7      & 98.0 $\pm$ 2.3          & 83.0 $\pm$ 6.0       & \textbf{47.0} $\pm$ 6.0      & 93.0 $\pm$ 3.8      & \textbf{100.0} $\pm$ 0.0       \\
\toprule
\textbf{Method}     & \textbf{Put in Cupboard} & \textbf{Put in Drawer} & \textbf{Put in Safe} & \textbf{Screw Bulb} & \textbf{Slide Block} & \textbf{Sort Shape} & \textbf{Stack Blocks} & \textbf{Stack Cups} & \textbf{Sweep to Dustpan} & \textbf{Turn Tap} \\
\midrule
PerAct \cite{shridhar2023perceiver}              & 28.0 $\pm$ 4.4           & 51.2 $\pm$ 4.7         & 84.0 $\pm$ 3.6       & 17.6 $\pm$ 2.0      & 74.0 $\pm$ 13.0      & 16.8 $\pm$ 4.7      & 26.4 $\pm$ 3.2        & 2.4 $\pm$ 2.0       & 52.0 $\pm$ 0.0            & 88.0 $\pm$ 4.4    \\
RVT \cite{goyal2023rvt}                 & 49.6 $\pm$ 3.2           & 88.0 $\pm$ 5.7         & 91.2 $\pm$ 3.0       & 48.0 $\pm$ 5.7      & 81.6 $\pm$ 5.4       & 36.0 $\pm$ 2.5      & 28.8 $\pm$ 3.9        & 26.4 $\pm$ 8.2      & 72.0 $\pm$ 0.0            & 93.6 $\pm$ 4.1    \\
RVT-2 \cite{goyal2024rvt}                & 66.0 $\pm$ 4.5           & 96.0 $\pm$ 0.0         & 96.0 $\pm$ 2.8       & 88.0 $\pm$ 4.9      & 92.0 $\pm$ 2.8       & 35.0 $\pm$ 7.1      & \textbf{80.0} $\pm$ 2.8        & 69.0 $\pm$ 5.9      & \textbf{100.0} $\pm$ 0.0           & 99.0 $\pm$ 1.7    \\
SAM-E \cite{zhang2024sam}               & 64.0 $\pm$ 2.8           & 92.0 $\pm$ 5.7         & 95.2 $\pm$ 3.3       & 78.4 $\pm$ 3.6      & \textbf{95.2}$\pm$ 1.8       & 34.4 $\pm$ 6.1      & 26.4 $\pm$ 4.6        & 0.0 $\pm$ 0.0       & \textbf{100.0} $\pm$ 0.0           & \textbf{100.0} $\pm$ 0.0   \\
SAM2Act (Ours)      & \textbf{75.0} $\pm$ 3.8           & \textbf{99.0} $\pm$ 2.0         & \textbf{98.0} $\pm$ 2.3       & \textbf{89.0} $\pm$ 2.0      & 86.0 $\pm$ 4.0       & \textbf{64.0} $\pm$ 4.6      & 76.0 $\pm$ 8.6        & \textbf{78.0} $\pm$ 4.0      & 99.0 $\pm$ 2.0            & 96.0 $\pm$ 5.7    \\

\bottomrule
\end{tabular}%
}
\label{tab:rlbench}
\end{table*}

\subsection{Semantic Generalization across Tasks}
\label{sec:experiments:generalization}

The results evaluated in \autoref{sec:experiments:baselines} were obtained by training and testing models within the same environment. However, to truly assess \textbf{generalization performance}, policies must remain robust against both environmental and object-level perturbations. We therefore trained SAM2Act and the baseline methods on 20 tasks from \texttt{The Colosseum} benchmark and tested them under 13 different perturbation categories over three runs. \textbf{SAM2Act exhibits the smallest performance drop compared to the baselines}, with an average decrease of 4.3\% (standard deviation of 3.59\%). Notably, it proves particularly robust to environmental perturbations -- such as changes in lighting, table color/texture, the addition of distractors, and even camera pose -- while also maintaining competitive performance under object-level perturbations (see more analysis in \autoref{app:sam2act_ablation:colosseum}).

\begin{table*}[h!]
\centering
\caption{\textbf{\texttt{The Colosseum} results}. Task-average success rate percentage change for SAM2Act and other baselines across 13 perturbation factors from \texttt{The Colosseum}, relative to evaluations without perturbations. Results of 3 evaluations are reported for all models. Our approach, SAM2Act, demonstrates the lowest average percentage change across all perturbations, with a minimal drop of -4.3$\pm$3.6\%, highlighting its robustness in handling various environmental and object-level perturbations. The full result table is shown in \autoref{app:colosseum_full}.}
\resizebox{\textwidth}{!}{%
\begin{tabular}{lcccccccccc}

\toprule
\textbf{Method}     & \textbf{Average $\uparrow$} & \textbf{MO-Color $\uparrow$} & \textbf{RO-Color $\uparrow$} & \textbf{MO-Texture $\uparrow$} & \textbf{RO-Texture $\uparrow$} & \textbf{MO-Size $\uparrow$} & \textbf{RO-Size $\uparrow$}  \\
\midrule
RVT-2 \cite{goyal2024rvt}            &        -19.5$\pm$2.8         &            -20.7$\pm$1.0                 &  -11.8$\pm$0.8  &   -13.3$\pm$4.6   &  -11.4$\pm$3.7&     \textbf{-13.2}$\pm$3.1     &  -17.7$\pm$0.1     &  \\
SAM2Act (SAM2 $\rightarrow$ SAM)            &     -20.7$\pm$1.2             &                   -26.1$\pm$0.7           &  -15.7$\pm$2.9 &    -15.0$\pm$3.3 &   -16.5$\pm$6.2    &   -18.7$\pm$1.9       &    -19.8$\pm$1.3    &  \\
SAM2Act (w/o Multi-res Input)            &         -19.1$\pm$4.5          &           -15.5$\pm$6.4              & -13.5$\pm$4.6  &  -20.4$\pm$0.5   & -16.6$\pm$6.1 &  -21.3$\pm$7.5     &  -12.6$\pm$7.5    &    \\
SAM2Act (Ours)               &       \textbf{-4.3}$\pm$3.6           &                     \textbf{-1.1}$\pm$2.5       &-\textbf{0.7}$\pm$7.2   &  \textbf{-3.3}$\pm$2.4    &  \textbf{24.72}$\pm$6.1    &    -15.9$\pm$5.0     &  \textbf{0.9}$\pm$6.8      &        \\

\toprule
\textbf{Method}     & \textbf{Light Color $\uparrow$} & \textbf{Table Color $\uparrow$} & \textbf{Table Texture $\uparrow$} & \textbf{Distractor $\uparrow$} & \textbf{Background Texture $\uparrow$} & \textbf{Camera Pose $\uparrow$} & \textbf{All Perturbations $\uparrow$}  \\
\midrule
RVT-2 \cite{goyal2024rvt}             &                            -15.6$\pm$1.3              &  -26.5$\pm$4.4  & -14.6$\pm$4.4   &   -4.9$\pm$5.3   &  -4.4$\pm$4.0     &      -19.5$\pm$2.8    &     -77.9$\pm$1.7    &   \\
SAM2Act (SAM2 $\rightarrow$ SAM)                &      -16.3$\pm$1.2            &           -23.5$\pm$5.3                   & -12.3$\pm$3.1  &   0.6$\pm$2.9  &   -5.4$\pm$3.2    &     -20.7$\pm$1.2   &      -79.5$\pm$2.5   &          \\
SAM2Act (w/o Multi-res Input)            &        -7.2$\pm$3.6           &              -18.3$\pm$6.1           &-17.5$\pm$3.3   & -4.6$\pm$3.5    & -5.7$\pm$3.5   &   -19.1$\pm$4.5    &   -73.8 $\pm$2.2   &    \\
SAM2Act (Ours)               &        \textbf{4.5}$\pm$4.4          &               \textbf{1.1}$\pm$2.5             &\textbf{-3.7}$\pm$5.2   &  \textbf{1.7}$\pm$1.7     &     \textbf{-1.5}$\pm$2.7   &   -\textbf{4.3}$\pm$3.6     &   \textbf{-58.3}$\pm$4.4     &        \\

\bottomrule
\end{tabular}%
}
\label{tab:colosseum}
\end{table*}

\subsection{Performance on \texttt{MemoryBench}}
\label{sec:experiments:memory}

In \autoref{tab:memorybench}, we evaluate SAM2Act+ against SoTA 3D BC model, RVT-2 on \texttt{MemoryBench}, training all models in a single-task setting to isolate memory-related challenges (e.g., opening the wrong drawer rather than unrelated mid-task failures). This setup ensures that performance differences stem from memory capabilities. For a random agent, the expected success rates are determined by the number of possible choices per task: 33\% for \texttt{reopen\_drawer} (three drawers), 25\% for \texttt{put\_block\_back} (four patches), and 50\% for \texttt{rearrange\_block} (two blocks). However, variations in task complexity, fixed training data, and imbalanced task distributions lead to slight deviations from these baselines. Our proposed memory-based model, SAM2Act+, demonstrates \textbf{a strong understanding of spatial memory}, achieving an average success rate of 94.3\% across all tasks. It \textbf{outperforms SAM2Act (without memory) by a huge margin of 39.3\% on \texttt{MemoryBench}}, highlighting the significant impact of explicit memory modeling.

\begin{table}[!htbp]
\centering
\caption{\textbf{Performance on \texttt{MemoryBench}.} We report the success rates for the three spatial memory tasks in \texttt{MemoryBench}. Our method, SAM2Act+, significantly outperforms all baseline methods that lack an explicit memory mechanism, achieving an average improvement of 37.6\% across all three tasks. Note that there is an update with \texttt{MemoryBench}, see more in \autoref{app:memorybench_update}.}
\resizebox{0.8\columnwidth}{!}{%
\begin{tabular}{@{}cccccc@{}}
\toprule
Methods / Tasks   &  Avg. Success $\uparrow$     & (a) Reopen Drawer & (b) Put Block Back &  (c) Rearrange Block  &    \\
\midrule     
RVT-2            &  54.0 $\pm$ 5.3  & 60.0 $\pm$ 0.0   & 50.0 $\pm$ 2.3 & 52.0 $\pm$ 3.3 \\
SAM2Act (Ours)   & 55.0 $\pm$  24.3    & 48.0 $\pm$ 0.0   & 35.0 $\pm$ 3.8  & 82.0 $\pm$ 2.3  \\
SAM2Act+ (Ours)   & \textbf{94.3 $\pm$ 9.0}   & \textbf{84.0} $\pm$ 0.0   & \textbf{100.0} $\pm$ 0.0  & \textbf{99.0} $\pm$ 2.0  \\
\bottomrule
\end{tabular}%

}
\label{tab:memorybench}
\end{table}

\subsection{Real-robot Evaluations}
\label{sec:experiments:real}

\autoref{tab:in-out-dist} presents our real-world experiment results, where our method achieves a 75\% task success rate, compared to 43\% for RVT-2. SAM2Act significantly outperforms the baseline in high-precision tasks (60\% vs 0\%). It excels in memory-based tasks, such as \texttt{(d) Push the same button}, which requires recalling the button’s previous location. Here, SAM2Act achieves 70\% success, while RVT-2, relying on random guessing, scores 40\%. We also test models' generalization against perturbations like lighting changes, distractors, and position variations. Additional details are in the \autoref{app:real_world}, with real-world rollout videos available on our project website.

\begin{table}[ht]
    \centering
    \caption{ \textbf{Real-world results.} We compare RVT2 against SAM2Act for the first three tasks and SAM2Act+ on the last real-world tasks (indicated with *), evaluating performance both in-distribution and out-of-distribution during test time. }
    \resizebox{0.7\textwidth}{!}{%
    \begin{tabular}{lcccc}
    \toprule
    & \multicolumn{2}{c}{\textbf{In-Distribution}} & \multicolumn{2}{c}{\textbf{Out-Distribution}} \\
    \cmidrule(lr){2-3}\cmidrule(lr){4-5}
    \textbf{Task} & \textbf{RVT-2} & \textbf{SAM2Act} & \textbf{RVT-2} & \textbf{SAM2Act} \\
    \midrule
    (a) turn on the lamp              & 0/10  & \textbf{6/10} & 0/10 & \textbf{6/10} \\
    (b) push button sequence & 4/10 & \textbf{9/10} & 1/10 & \textbf{9/10} \\
    (c) stack cubes     & 8/10  & 8/10 & 3/10 & 3/10 \\
    (d) push the same button *  & 4/10 & \textbf{7/10} & 2/10 & \textbf{6/10} \\
    \bottomrule
    \end{tabular}
    }
    
    \label{tab:in-out-dist}
\end{table}

\section{Conclusion \& Limitation}
\label{sec:conclusion}

We introduce SAM2Act, a multi-view, language-conditioned behavior cloning policy for 6-DoF 3D manipulation, enabling high-precision manipulations while generalizing effectively to unseen perturbations. Building on this foundation, we propose SAM2Act+, a memory-based multi-view language-conditioned robotic transformer-based policy that equips the agent with spatial memory awareness, allowing it to solve spatial memory-based tasks. While both SAM2Act and SAM2Act+ achieve SOTA performance across multiple benchmarks, challenges remain in extending them to dexterous continuous control. Additionally, SAM2Act+ relies on a fixed memory window length, which differs from task to task, limiting its adaptability to tasks of varying length. We also examined whether our memory architecture could retain semantic information (e.g., color), but unfortunately, it appears to be limited to storing spatial information. Despite these challenges, we believe SAM2Act+ is an important step towards memory-based generalist manipulation policies.

\section{Acknowledgement}
Jiafei Duan is supported by the Agency for Science, Technology and Research (A*STAR) National Science Fellowship. Wilbert Pumacay is supported by grant 234-2015-FONDECYT from Cienciactiva of the National Council for Science, Technology and Technological Innovation (CONCYTEC-PERU). This project is partially supported by Amazon Science. We would also like to thank Winson Han from the Allen Institute for Artificial Intelligence for helping with the figure and icon design, and Jieyu Zhang from the University of Washington for assisting with the design of model architecture and training pipeline.

\bibliography{main}

\newpage

\appendix
\section{Model Architecture}
\label{app:architecture}

We will explain our model architecture in detail, including Multi-View Transformer, Memory Attention, Memory Encoder, and Memory Bank. The multi-resolution is already explained in \autoref{sec:method:sam2act}.

\textbf{Multi-View Transformer.} The two MVTs used in the coarse and fine branches have the same architecture. Very similar to the MVT proposed by \cite{goyal2023rvt}, the input to the transformer consists of a language description of the task, virtual images of the scene point cloud, and the image embeddings (at the lowest resolution) generated by the SAM2 image encoder. The text is transformed into token embeddings using the pre-trained CLIP \cite{radford2021clip} model, while the virtual images are converted into token embeddings through patchify and projection operations. Similarly, the image embeddings are converted into token embeddings via a projection layer. For each virtual image, tokens corresponding to the same image are processed through four attention layers. Finally, the processed image tokens, along with the language tokens, are jointly processed using an additional four attention layers. The resulting image tokens are then used to infer the 3D action. 

\textbf{Memory Attention.} Akin to the memory attention in SAM2 \cite{ravi2024sam}, the purpose of this module is to condition the current observation features on both past observation features and predicted actions, specifically translation. Notably, features from each view are processed independently. We stack four transformer blocks, with the first one taking the image embedding output of MVT from the current observation as input. Each block applies self-attention, followed by cross-attention to memories of past observation features and predicted actions, stored in a memory bank (described below), and ends with a multi-layer perceptron (MLP). For both self- and cross-attention, we use vanilla attention operations, enabling us to leverage recent advances in efficient attention kernels \cite{dao2023flashattn2}. In addition to sinusoidal absolute positional embeddings, 2D spatial Rotary Positional Embedding (RoPE) \cite{su2021roformer, heo2024rope} are incorporated in both self-attention and cross-attention layers. We also reduce the dimension size from the original 256 to 128 to align with the image embedding dimension of the MVT output.

\textbf{Memory Encoder.} The memory encoder constructs memory features by downsampling the output translation heatmap using a convolutional module and summing it element-wise with the unconditioned observation embedding from the multi-view transformer (not shown in \autoref{fig:sam2act}). This is followed by lightweight convolutional layers to integrate the information. Instead of employing an additional image encoder, our memory encoder reuses the image embeddings produced by the MVT (not the SAM2 image encoder) and fuses them with the predicted translation information to generate memory features. This design enables the memory features to leverage rich representations that incorporate language, semantic, and spatial features from multiple views, making them more suitable for encoding action memories. Originally, this module was designed to encode an image embedding with multiple object masks within the same frame. However, we do not utilize this functionality. Instead, we encode one memory per view, where each memory is generated by encoding a single heatmap with a corresponding image embedding from each view.

\textbf{Memory Bank.} The memory bank preserves past translation predictions associated with previous observations in the video by maintaining a FIFO queue of up to $N$ recent memories. Each view has its own independent memory bank, as memories are stored and retrieved separately for different views. These memories are represented as spatial feature maps. Additionally, in our memory bank, the memory features are projected to a dimension of 64.

\section{Training Implementation}
\label{app:training}

All models are trained on 32 NVIDIA H100/A100 GPUs. In some cases, we also train on 16 or 8 NVIDIA H100/A100 GPUs, but we ensure fairness by maintaining the same total batch size across all settings.

\subsection{SAM2Act}

We use the same way to data and demo augmentation methods and training pipeline as in RVT2 \cite{goyal2024rvt} to train SAM2Act (stage 1). The training hyperparameters are shown in \autoref{appendix_tab:hyperparam_rlbench}. We use this set of hyperparameters to train on RLBench and \texttt{The Colosseum}.

\begin{table}[!htbp]
\centering
\caption{\textbf{Training Hyperparameters of SAM2Act on RLBench and \texttt{The Colosseum}.} The batch size stands for total batch size across all GPUs. For the learning rate, we follow the scaling strategy used in RVT2 \cite{goyal2024rvt}, where the learning rate is scaled by the batch size as $1.25e-5 \times bs$.}
\vspace{0.2em}
\resizebox{0.4\textwidth}{!}{%
\begin{tabular}{@{}ccc@{}}
\toprule
Hyperparameters & SAM2Act Training    \\
\midrule     
batch size        &256                \\
learning rate        &3.2e-3            \\
optimizer          &LAMB            \\
learning rate schedule  &cosine decay  \\
weight decay       &1e-4              \\
warmup steps        &2000             \\
training steps     & 56.25K              \\
training epochs     &90                   \\
LoRA rank          &16                  \\
\bottomrule
\end{tabular}
}
\label{appendix_tab:hyperparam_rlbench}
\end{table}

\subsection{SAM2Act+}

We use a different strategy for sampling a batch of data for training. Previous sampling strategies randomly select a batch of independent observations, allowing the model to predict the next action based on each observation independently. However, for SAM2Act+, we aim for the agent to predict the next action based on both the current and past observations. To achieve this, we must sample a batch of data that is spatio-temporally consistent. To implement this, we randomly sample $n$ consecutive observations from a random episode. The forward pass is then performed sequentially from the first to the last observation. The details of the forward pass are provided in \autoref{algo:forward}.

When adopting this new sampling method during training, one immediate effect is a significant reduction in data diversity per batch. This can be detrimental, especially when dealing with tasks with numerous variations. We attempted to train the standard SAM2Act model on RLBench tasks using this new sampling method, but the convergence time was excessively long. To address this, we propose a new training pipeline: first, we pre-train the model using the previous sampling method, then fine-tune it with the new sampling approach. This strategy effectively mitigates the issue of slow convergence, significantly reducing training time.

As mentioned in \autoref{sec:experiments:memory}, we train all methods on \texttt{MemoryBench} in a single-task setting. However, finding a training configuration that optimizes all tasks is challenging. To address this, we use a universal set of hyperparameters for training but evaluate models across all epochs and select the best-performing one for evaluation. We follow the same approach to determine the optimal pre-trained weights for SAM2Act before fine-tuning on SAM2Act+. In addition, the window size of the memory mechanism is also decided to be different for each task in \texttt{MemoryBench}. We keep the batch size the same as the window size during training, and thus the learning rate will be a bit different as they are related with batch size. The detailed training hyperparameters are listed in \autoref{appendix_tab:hyperparam_memory}.

\begin{table}[!htbp]
\centering
\caption{\textbf{Training Hyperparameters of SAM2Act and SAM2Act+ on \texttt{MemoryBench}.} Note that the batch size refers to the total batch size across all GPUs. For SAM2Act+ training on the \texttt{reopen\_drawer} task, we use a maximum window size of 8, resulting in a per-GPU batch size of 8 and a total batch size of 256. Similarly, for the other two tasks, where the maximum window size is 10, the total batch size is $10 \times 32=320$ in total. The learning rate follows the same scaling rule mentioned in \autoref{appendix_tab:hyperparam_rlbench}.}
\vspace{0.2em}
\resizebox{0.8\textwidth}{!}{%
\begin{tabular}{@{}cccc@{}}
\toprule
Hyperparameters & SAM2Act Training  &  SAM2Act+ Training\\
\midrule     
batch size        &256        & 256 ($\texttt{reopen\_drawer}$), 320 (other two)      \\
learning rate      &3.2e-3        & 3.2e-3 ($\texttt{reopen\_drawer}$), 4e-3 (other two)       \\
optimizer          &LAMB          & LAMB       \\
learning rate schedule  &cosine decay   & cosine decay  \\
weight decay       &1e-4          & 1e-4       \\
warmup steps        &2000          & 2000      \\
training steps     & 6.25K        & 12.5K        \\
training epochs     &10             & 20      \\
LoRA rank          &16              & 16         \\
\bottomrule
\end{tabular}
}
\label{appendix_tab:hyperparam_memory}
\end{table}


\section{Full Comparisons for RLBench 18 Tasks}
\label{app:rlbench_full}

The full comparisons of SAM2Act with existing approaches on RLBench 18 tasks are shown in \autoref{appendix_tab:rlbench_full}.

\begin{table*}[!h]
\centering
\caption{\textbf{Full Comparisons of Multi-Task Performance on RLBench.} We evaluate 18 RLBench tasks \cite{james2020rlbench}, reporting success rates across all tasks among all existing approaches, not limited to 3D keyframe-based behavior cloning (BC) policies. We report stats of 4 evaluations for SAM2Act. Our method, SAM2Act, outperforms all baselines, achieving a performance margin of 1.9\% over ARP$^+$ \cite{zhang2024arp}, the prior state-of-the-art approach.}
\resizebox{\textwidth}{!}{%
\begin{tabular}{lccccccccccccc}

\toprule
\textbf{Method}     & \textbf{Avg. Success $\uparrow$} & \textbf{Avg. Rank $\downarrow$} & \textbf{Close Jar}   & \textbf{Drag Stick}        & \textbf{Insert Peg}     & \textbf{Meat off Grill}   & \textbf{Open Drawer}       & \textbf{Place Cups}      & \textbf{Place Wine}      & \textbf{Push Buttons} \\
\midrule
Image-BC (CNN) \cite{jang2022bcz}     & 1.3                     & 12.4                    & 0.0                      & 0.0                      & 0.0                      & 0.0                     & 4.0                     & 0.0                      & 0.0                       & 0.0 \\
Image-BC (ViT) \cite{jang2022bcz}     & 1.3                     & 12.6                    & 0.0                      & 0.0                      & 0.0                      & 0.0                     & 0.0                     & 0.0                      & 0.0                       & 0.0 \\
C2F-ARM-BC \cite{james2021c2f}        & 20.1                    & 11.5                    & 24.0                     & 24.0                     & 4.0                      & 20.0                    & 20.0                    & 0.0                      & 8.0                       & 72.0 \\
HiveFormer \cite{guhur2022hiveformer} & 45.3                    & 9.6                     & 52.0                     & 76.0                     & 0.0                      & \textbf{100.0}          & 52.0                    & 0.0                      & 80.0                      & 84.0 \\
PolarNet \cite{chen2023polarnet}      & 46.4                    & 9.1                     & 36.0                     & 92.0                     & 4.0                      & \textbf{100.0}          & 84.0                    & 0.0                      & 40.0                      & 96.0 \\
PerAct \cite{shridhar2023perceiver}   & 49.4 $\pm$ 4.3          & 8.9                     & 55.2 $\pm$ 4.7           & 89.6 $\pm$ 4.1           & 5.6 $\pm$ 4.1            & 70.4 $\pm$ 2.0          & 88.0 $\pm$ 5.7          & 2.4 $\pm$ 3.2            & 44.8 $\pm$ 7.8            & 92.8 $\pm$ 3.0        \\
Act3D \cite{gervet2023act3d}          & 65.0                    & 6.8                     & 92.0                     & 92.0                     & 27.0                     & 94.0                    & 93.0                    & 3.0                      & 80.0                      & 99.0 \\
RVT \cite{goyal2023rvt}               & 62.9 $\pm$ 3.7          & 6.9                     & 52.0 $\pm$ 2.5           & 99.2 $\pm$ 1.6           & 11.2 $\pm$ 3.0           & 88.0 $\pm$ 2.5          & 71.2 $\pm$ 6.9          & 4.0 $\pm$ 2.5            & 91.0 $\pm$ 5.2            & \textbf{100.0} $\pm$ 0.0       \\
RVT-2 \cite{goyal2024rvt}             & 81.4 $\pm$ 3.1          & 3.7                     & \textbf{100.0} $\pm$ 0.0 & 99.0 $\pm$ 1.7           & 40.0 $\pm$ 0.0           & 99.0 $\pm$ 1.7          & 74.0 $\pm$ 11.8         & 38.0 $\pm$ 4.5           & 95.0 $\pm$ 3.3            & \textbf{100.0} $\pm$ 0.0       \\
3D Diffuser Actor \cite{ke20243dda}   & 81.3                    & 3.9                     & 96.0 $\pm$ 2.5           & \textbf{100.0} $\pm$ 0.0 & 65.6 $\pm$ 4.1           & 96.8 $\pm$ 1.6          & 89.6 $\pm$ 4.1          & 24.0 $\pm$ 7.6           & 93.6 $\pm$ 4.8            & 98.4 $\pm$ 2.0              \\
3D-LOTUS \cite{garcia20243dlotus}     & 83.1                    & 3.7                     & 96.0 $\pm$ 0.0           & \textbf{100.0} $\pm$ 0.0 & 69.6 $\pm$ 3.6           & 98.4 $\pm$ 2.2          & 85.6 $\pm$ 7.3          & 40.8 $\pm$ 12.1          & 91.2 $\pm$ 6.6            & \textbf{100.0} $\pm$ 0.0    \\
ARP$^+$ \cite{zhang2024arp}           & 84.9                    & 3.2                     & 95.2                     & 99.2                     & 78.4                     & 97.6                    & 92.8                    & \textbf{48.8}            & \textbf{96.0}             & \textbf{100.0}              \\
SAM-E \cite{zhang2024sam}             & 70.6 $\pm$ 0.7          & 4.8                     & 82.4 $\pm$ 3.6           & \textbf{100.0} $\pm$ 0.0 & 18.4 $\pm$ 4.6           & 95.2 $\pm$ 3.3          & \textbf{95.2} $\pm$ 5.2 & 0.0 $\pm$ 0.0            & 94.4 $\pm$ 4.6            & \textbf{100.0} $\pm$ 0.0       \\
SAM2Act (Ours)                        & \textbf{86.8} $\pm$ 0.5 & \textbf{3.1}            & 99.0 $\pm$ 2.0           & 99.0 $\pm$ 2.0           & \textbf{84.0} $\pm$ 5.7  & 98.0 $\pm$ 2.3          & 83.0 $\pm$ 6.0          & 47.0 $\pm$ 6.0           & 93.0 $\pm$ 3.8            & \textbf{100.0} $\pm$ 0.0       \\
\toprule
\textbf{Method}                     & \textbf{Put in Cupboard} & \textbf{Put in Drawer} & \textbf{Put in Safe}   & \textbf{Screw Bulb}   & \textbf{Slide Block}     & \textbf{Sort Shape}       & \textbf{Stack Blocks}      & \textbf{Stack Cups}   & \textbf{Sweep to Dustpan}     & \textbf{Turn Tap} \\
\midrule
Image-BC (CNN) \cite{jang2022bcz}     & 0.0                     & 8.0                     & 4.0                      & 0.0                      & 0.0                      & 0.0                     & 0.0                     & 0.0                      & 0.0                       & 8.0  \\
Image-BC (ViT) \cite{jang2022bcz}     & 0.0                     & 0.0                     & 0.0                      & 0.0                      & 0.0                      & 0.0                     & 0.0                     & 0.0                      & 0.0                       & 16.0 \\
C2F-ARM-BC \cite{james2021c2f}        & 0.0                     & 4.0                     & 12.0                     & 8.0                      & 16.0                     & 8.0                     & 0.0                     & 0.0                      & 0.0                       & 68.0 \\
HiveFormer \cite{guhur2022hiveformer} & 32.0                    & 68.0                    & 76.0                     & 8.0                      & 64.0                     & 8.0                     & 8.0                     & 0.0                      & 28.0                      & 80.0  \\
PolarNet \cite{chen2023polarnet}      & 12.0                    & 32.0                    & 84.0                     & 44.0                     & 56.0                     & 12.0                    & 4.0                     & 8.0                      & 52.0                      & 80.0  \\
PerAct \cite{shridhar2023perceiver}   & 28.0 $\pm$ 4.4          & 51.2 $\pm$ 4.7          & 84.0 $\pm$ 3.6           & 17.6 $\pm$ 2.0           & 74.0 $\pm$ 13.0          & 16.8 $\pm$ 4.7          & 26.4 $\pm$ 3.2          & 2.4 $\pm$ 2.0            & 52.0 $\pm$ 0.0            & 88.0 $\pm$ 4.4    \\
Act3D \cite{gervet2023act3d}          & 51.0                    & 90.0                    & 95.0                     & 47.0                     & 93.0                     & 8.0                     & 12.0                    & 9.0                      & 92.0                      & 94.0                  \\
RVT \cite{goyal2023rvt}               & 49.6 $\pm$ 3.2          & 88.0 $\pm$ 5.7          & 91.2 $\pm$ 3.0           & 48.0 $\pm$ 5.7           & 81.6 $\pm$ 5.4           & 36.0 $\pm$ 2.5          & 28.8 $\pm$ 3.9          & 26.4 $\pm$ 8.2           & 72.0 $\pm$ 0.0            & 93.6 $\pm$ 4.1    \\
RVT-2 \cite{goyal2024rvt}             & 66.0 $\pm$ 4.5          & 96.0 $\pm$ 0.0          & 96.0 $\pm$ 2.8           & 88.0 $\pm$ 4.9           & 92.0 $\pm$ 2.8           & 35.0 $\pm$ 7.1          & \textbf{80.0} $\pm$ 2.8 & 69.0 $\pm$ 5.9           & \textbf{100.0} $\pm$ 0.0  & 99.0 $\pm$ 1.7    \\
3D Diffuser Actor \cite{ke20243dda}   & \textbf{85.6} $\pm$ 4.1 & 96.0 $\pm$ 3.6          & 97.6 $\pm$ 2.0           & 82.4 $\pm$ 2.0           & 97.6 $\pm$ 3.2           & 44.0 $\pm$ 4.4          & 68.3 $\pm$ 3.3          & 47.2 $\pm$ 8.5           & 84.0 $\pm$ 4.4            & 99.2 $\pm$ 1.6      \\
3D-LOTUS \cite{garcia20243dlotus}     & 78.4 $\pm$ 4.6          & 97.6 $\pm$ 3.6          & 95.2 $\pm$ 3.4           & 88.8 $\pm$ 3.4           & \textbf{99.2} $\pm$ 1.8  & 34.4 $\pm$ 4.6          & 58.4 $\pm$ 8.3          & 75.2 $\pm$ 7.7           & 96.0 $\pm$ 2.8            & 90.4 $\pm$ 4.6      \\
ARP$^+$ \cite{zhang2024arp}           & 69.6                    & 98.4                    & 86.4                     & \textbf{89.6}            & 92.8                     & 46.4                    & 63.2                    & \textbf{80.0}            & 97.6                      & 96.0                \\
SAM-E \cite{zhang2024sam}             & 64.0 $\pm$ 2.8          & 92.0 $\pm$ 5.7          & 95.2 $\pm$ 3.3           & 78.4 $\pm$ 3.6           & 95.2 $\pm$ 1.8           & 34.4 $\pm$ 6.1          & 26.4 $\pm$ 4.6          & 0.0 $\pm$ 0.0            & \textbf{100.0} $\pm$ 0.0  & \textbf{100.0} $\pm$ 0.0   \\
SAM2Act (Ours)                        & 75.0 $\pm$ 3.8          & \textbf{99.0} $\pm$ 2.0 & \textbf{98.0} $\pm$ 2.3  & 89.0 $\pm$ 2.0           & 86.0 $\pm$ 4.0           & \textbf{64.0} $\pm$ 4.6 & 76.0 $\pm$ 8.6          & 78.0 $\pm$ 4.0           & 99.0 $\pm$ 2.0            & 96.0 $\pm$ 5.7           \\

\bottomrule
\end{tabular}%
}
\label{appendix_tab:rlbench_full}
\end{table*}

\section{Full Results for \texttt{The Colosseum}}
\label{app:colosseum_full}

The full results of SAM2Act on \texttt{The Colosseum} are shown in \autoref{appendix_tab:colosseum_full}.

\begin{table*}[!h]
    \centering
    \caption{\textbf{Full Results of SAM2Act for Various Perturbations on \texttt{The Colosseum}.} Mean and std of 3 evaluations are reported.}
    \resizebox{\textwidth}{!}{%
    \begin{tabular}{lcccccccccccccc}
    \toprule
    \textbf{Task Name} &
    \textbf{No Variations} &
    \textbf{All Variations} &
    \textbf{MO Color} &
    \textbf{RO Color} &
    \textbf{MO Texture} &
    \textbf{RO Texture} &
    \textbf{MO Size} &
    \textbf{RO Size} &
    \textbf{Light Color} &
    \textbf{Table Color} &
    \textbf{Table Texture} &
    \textbf{Distractor} &
    \textbf{Background Texture} &
    \textbf{Camera Pose} \\
    \midrule
    basketball\_in\_hoop            & 100.0 $\pm$ 0.0 & 30.7 $\pm$ 2.3 & 97.3 $\pm$ 2.3 & 100.0 $\pm$ 0.0 & 97.3 $\pm$ 2.3 & –              & 100.0 $\pm$ 0.0 & 86.7 $\pm$ 2.3 & 100.0 $\pm$ 0.0 & 100.0 $\pm$ 0.0 & 100.0 $\pm$ 0.0 & 94.7 $\pm$ 4.6 & 100.0 $\pm$ 0.0 & 100.0 $\pm$ 0.0 \\
    close\_box                      &  89.3 $\pm$ 6.1 & 61.3 $\pm$ 6.1 & 85.3 $\pm$ 6.1 & –               & –               & –              &  90.7 $\pm$ 6.1 & –               &  90.7 $\pm$ 2.3 &  85.3 $\pm$ 2.3 &  81.3 $\pm$ 2.3 & 93.3 $\pm$ 4.6 &  97.3 $\pm$ 4.6 &  92.0 $\pm$ 6.9 \\
    close\_laptop\_lid              &  96.0 $\pm$ 0.0 & 60.0 $\pm$ 0.0 & 100.0 $\pm$ 0.0& –               & –               & –              &  93.3 $\pm$ 11.5& –               &  94.7 $\pm$ 4.6 &  96.0 $\pm$ 0.0 &  84.0 $\pm$ 0.0 & 93.3 $\pm$ 2.3 &  96.0 $\pm$ 0.0 &  96.0 $\pm$ 0.0 \\
    empty\_dishwasher               &   0.0 $\pm$ 0.0 &  1.3 $\pm$ 2.3 &  0.0 $\pm$ 0.0 &  0.0 $\pm$ 0.0 & –               & 0.0 $\pm$ 0.0  &   1.3 $\pm$ 2.3 &  0.0 $\pm$ 0.0 &   0.0 $\pm$ 0.0 &   0.0 $\pm$ 0.0 &   0.0 $\pm$ 0.0 &  0.0 $\pm$ 0.0 &   0.0 $\pm$ 0.0 &   0.0 $\pm$ 0.0 \\
    get\_ice\_from\_fridge          &  93.3 $\pm$ 4.6 & 41.3 $\pm$ 2.3 & 92.0 $\pm$ 0.0 & 93.3 $\pm$ 2.3 & 89.3 $\pm$ 2.3 & –              &  84.0 $\pm$ 6.9 & 81.3 $\pm$ 2.3 &  85.3 $\pm$ 9.2 &  98.7 $\pm$ 2.3 &  94.7 $\pm$ 2.3 & 93.3 $\pm$ 2.3 &  89.3 $\pm$ 2.3 & 100.0 $\pm$ 0.0 \\
    hockey                          &  16.0 $\pm$ 4.0 &  0.0 $\pm$ 0.0 & 30.7 $\pm$ 2.3 & 14.7 $\pm$ 2.3 & –               &  9.3 $\pm$ 4.6 &  18.7 $\pm$ 4.6 & 21.3 $\pm$ 2.3 &  29.3 $\pm$ 2.3 &  52.0 $\pm$ 6.9 &  26.7 $\pm$ 4.6 &  6.7 $\pm$ 4.6 &  21.3 $\pm$ 2.3 &  40.0 $\pm$ 6.9 \\
    meat\_on\_grill                 &  98.7 $\pm$ 2.3 & 34.7 $\pm$ 2.3 &100.0 $\pm$ 0.0 &100.0 $\pm$ 0.0 & –               & –              &  98.7 $\pm$ 2.3 & –               &  62.7 $\pm$ 28.9&  69.3 $\pm$ 9.2 &  76.0 $\pm$ 4.0 &100.0 $\pm$ 0.0 &  98.7 $\pm$ 2.3 &  98.7 $\pm$ 2.3 \\
    move\_hanger                    &   1.3 $\pm$ 2.3 & 12.0 $\pm$ 0.0 & 32.0 $\pm$ 0.0 &  0.0 $\pm$ 0.0 & –               & –              & –               & –               &  49.3 $\pm$ 2.3 &  64.0 $\pm$ 0.0 &  44.0 $\pm$ 6.9 & 36.0 $\pm$ 6.9 &   0.0 $\pm$ 0.0 &  37.3 $\pm$ 18.5\\
    wipe\_desk                      &   0.0 $\pm$ 0.0 &  0.0 $\pm$ 0.0 &  0.0 $\pm$ 0.0 & –               &  0.0 $\pm$ 0.0 & –              &  0.0 $\pm$ 0.0 & –               &   0.0 $\pm$ 0.0 &   0.0 $\pm$ 0.0 &   0.0 $\pm$ 0.0 &  0.0 $\pm$ 0.0 &   1.3 $\pm$ 2.3 &   0.0 $\pm$ 0.0 \\
    open\_drawer                    &  94.7 $\pm$ 2.3 & 70.7 $\pm$ 6.1 & 96.0 $\pm$ 0.0 & –               & –               & –              &  92.0 $\pm$ 0.0 & –               &  88.0 $\pm$ 0.0 &  88.0 $\pm$ 0.0 & 100.0 $\pm$ 0.0 & 85.3 $\pm$ 2.3 &  98.7 $\pm$ 2.3 &  90.7 $\pm$ 4.6 \\
    slide\_block\_to\_target        &  12.0 $\pm$ 0.0 & 29.3 $\pm$ 11.5& 42.7 $\pm$ 4.6 & –               & 25.3 $\pm$ 4.6 & –              & –               & –               &  25.3 $\pm$ 4.6 &  40.0 $\pm$ 0.0 &  90.7 $\pm$ 2.3 & 49.3 $\pm$ 9.2 &  18.7 $\pm$ 2.3 &  24.0 $\pm$ 0.0 \\
    reach\_and\_drag                &  65.3 $\pm$ 14.0&  1.3 $\pm$ 2.3 & 54.7 $\pm$ 4.6 & 80.0 $\pm$ 10.6& 51.7 $\pm$ 4.0 & 69.3 $\pm$ 2.3&  52.0 $\pm$ 27.7& 37.3 $\pm$ 2.3 &  76.0 $\pm$ 6.9 &  81.3 $\pm$ 12.2&  70.7 $\pm$ 26.6& 65.3 $\pm$ 8.3 &  70.7 $\pm$ 16.2&  58.7 $\pm$ 16.2\\
    put\_money\_in\_safe            &  74.7 $\pm$ 2.3 & 20.0 $\pm$ 4.0 & 54.7 $\pm$ 8.3 & 52.0 $\pm$ 10.6& 37.3 $\pm$ 2.3 & 66.7 $\pm$ 14.0&  69.3 $\pm$ 2.3 & –               &  73.3 $\pm$ 11.5&  69.3 $\pm$ 2.3 &  76.0 $\pm$ 20.8& 77.3 $\pm$ 14.0&  50.7 $\pm$ 23.4&  45.3 $\pm$ 18.5\\
    place\_wine\_at\_rack\_location &  98.7 $\pm$ 2.3 & 38.7 $\pm$ 4.6 & 81.3 $\pm$ 2.3 & 85.3 $\pm$ 2.3 & –               & 96.0 $\pm$ 6.9 &  90.7 $\pm$ 4.6 & 97.3 $\pm$ 4.6 &  86.7 $\pm$ 4.6 &  88.0 $\pm$ 0.0 &  97.3 $\pm$ 4.6 & 86.7 $\pm$ 4.6 &  92.0 $\pm$ 6.9 &  69.3 $\pm$ 39.3\\
    insert\_onto\_square\_peg       &  88.0 $\pm$ 6.9 & 46.7 $\pm$ 39.3& 60.0 $\pm$ 4.0 & 98.7 $\pm$ 2.3 & –               & 69.3 $\pm$ 4.6 &  58.7 $\pm$ 2.3 & 61.3 $\pm$ 6.1 &  80.0 $\pm$ 0.0 &  82.7 $\pm$ 4.6 &  64.0 $\pm$ 4.0 & 58.7 $\pm$ 2.3 &  90.7 $\pm$ 2.3 &  82.7 $\pm$ 2.3 \\
    stack\_cups                     &  89.3 $\pm$ 4.6 &  1.3 $\pm$ 2.3 & 88.0 $\pm$ 0.0 & –               & 78.7 $\pm$ 2.3 & –              &  53.3 $\pm$ 11.5& –               &  88.0 $\pm$ 0.0 &  61.3 $\pm$ 2.3 &  46.7 $\pm$ 2.3 & 73.3 $\pm$ 25.4&  81.3 $\pm$ 2.3 &  84.0 $\pm$ 0.0 \\
    turn\_oven\_on                  &  96.0 $\pm$ 0.0 & 72.0 $\pm$ 0.0 & 92.0 $\pm$ 6.9 & –               & –               & –              &  88.0 $\pm$ 0.0 & –               &  89.3 $\pm$ 4.6 &  96.0 $\pm$ 0.0 &  98.7 $\pm$ 2.3 & 96.0 $\pm$ 0.0 &  96.0 $\pm$ 0.0 &  98.7 $\pm$ 2.3 \\
    straighten\_rope                &  78.7 $\pm$ 9.2 &  6.7 $\pm$ 2.3 & 65.3 $\pm$ 4.6 & –              & 70.7 $\pm$ 4.6 & –              & –               & –               &  84.0 $\pm$ 0.0 &  64.0 $\pm$ 0.0 &  61.3 $\pm$ 2.3 & 49.3 $\pm$ 2.3 &  90.7 $\pm$ 9.2 &  76.0 $\pm$ 0.0 \\
    setup\_chess                    &  10.7 $\pm$ 2.3 &  0.0 $\pm$ 0.0 & 12.0 $\pm$ 0.0 & 18.7 $\pm$ 2.3 & 16.0 $\pm$ 0.0 & –              &  26.7 $\pm$ 2.3 & –               &  22.7 $\pm$ 2.3 &  34.7 $\pm$ 11.5&  20.0 $\pm$ 6.9 & 22.7 $\pm$ 4.6 &  28.0 $\pm$ 6.9 &  26.7 $\pm$ 2.3 \\
    scoop\_with\_spatula            &  92.0 $\pm$ 6.9 & 10.7 $\pm$ 2.3 & 96.0 $\pm$ 6.9 & 89.3 $\pm$ 2.3 & 88.0 $\pm$ 6.9 & 92.0 $\pm$ 6.9 &  94.7 $\pm$ 9.2 & 78.7 $\pm$ 2.3 &  78.7 $\pm$ 4.6 &  81.3 $\pm$ 4.6 &  76.0 $\pm$ 6.9 & 64.0 $\pm$ 6.9 &  96.0 $\pm$ 0.0 &  94.7 $\pm$ 4.6 \\
    \bottomrule
    \end{tabular}
    }
    \label{appendix_tab:colosseum_full}
\end{table*}

\section{Ablation on SAM2Act}
\label{app:sam2act_ablation}

\subsection{RLBench}
\label{app:sam2act_ablation:rlbench}

We conduct ablation experiments on the proposed SAM2Act, focusing on two key aspects: the SAM2 image Encoder and multi-resolution upsampling. We evaluate the model under three different configurations:

(i) Replacing the SAM2 image encoder with the SAM image encoder and removing the multi-resolution upsampling, as the SAM image encoder does not produce multi-resolution outputs.
(ii) Replacing the multi-resolution upsampling with the original convex upsampling from RVT-2 \cite{goyal2024rvt}.
(iii) Removing SAM2's multi-resolution image embedding inputs to the multi-resolution upsampling while keeping the multi-resolution upsampling itself.

Note that SAM-E \cite{zhang2024sam} proposed a 3D behavior cloning policy that integrates RVT and the SAM image encoder, along with an action-sequence policy head. We attempted to extend this method to the more powerful RVT2 backbone for comparison. However, its action-sequence policy proved incompatible with the coarse-to-fine pipeline, resulting in very slow convergence under SAM-E’s training setup. To ensure a fair comparison, we also extended SAM-E while keeping its original hyperparameters (notably, a LoRA rank of 4, whereas ours is 16). We trained both versions and found that SAM-E’s configuration performed better. Therefore, we adopted their configuration and reported the results accordingly, which also applies to \autoref{sec:experiments:generalization}. For all other ablation experiments, the training configuration are kept the same. 

Ablation results on RLBench are presented in \autoref{appendix_tab:rlbench_abl}. All variants of SAM2Act exhibit lower performance than the original version. Removing SAM2’s multi-resolution image embedding inputs results in a 1.1\% drop in the average success rate. Replacing the entire multi-resolution upsampling with the original convex upsampling leads to a 2.6\% decrease. Substituting the SAM2 image encoder with the SAM image encoder \cite{kirillov2023sam} causes a 6.0\% drop compared to SAM2Act and a 3.4\% drop compared to SAM2Act with the original convex upsampling, where the only differences are the image encoder and some training hyperparameters. In the same setting, we further replace the SAM2 image encoder to latest image encoders, DINOv2 \cite{oquab2023dinov2} and Depth Anything V2 \cite{yang2024dav2}, while both of them show a large drop compared to the original SAM2Act. These results indicate that all of our architectural innovations significantly enhance the agent’s ability across multiple manipulation tasks.

\begin{table*}[h!]
\centering
\caption{\textbf{SAM2Act Abaltion Performance on RLBench.} We report the success rates for 18 RLBench tasks \cite{james2020rlbench}, along with the average success rate and ranking across all tasks. Table shows that SAM2Act outperforms all of its variations.}
\resizebox{\textwidth}{!}{%
\begin{tabular}{lccccccccccccc}

\toprule
\textbf{Method}                     & \textbf{Avg. Success $\uparrow$} & \textbf{Avg. Rank $\downarrow$} & \textbf{Close Jar} & \textbf{Drag Stick} & \textbf{Insert Peg} & \textbf{Meat off Grill} & \textbf{Open Drawer} & \textbf{Place Cups} & \textbf{Place Wine} & \textbf{Push Buttons} \\
\midrule
SAM2Act (SAM2 $\rightarrow$ SAM)    & 80.8 $\pm$ 1.9                   & 3.9                         & 96.0 $\pm$ 3.3          & 94.0 $\pm$ 4.0      & 28.0 $\pm$ 8.6      & 98.0 $\pm$ 2.3          & 72.0 $\pm$ 7.3       & 42.0 $\pm$ 6.9      & 95.0 $\pm$ 3.8      & \textbf{100.0} $\pm$ 0.0        \\
SAM2Act (SAM2 $\rightarrow$ Depth Anything V2) & 81.1 $\pm$ 1.2          & 3.6                         &\textbf{100.0} $\pm$ 0.0 & 98.0 $\pm$ 2.3      & 58.0 $\pm$ 6.9      & \textbf{99.0} $\pm$ 2.0 & 81.0 $\pm$ 3.8       & 24.0 $\pm$ 8.6      & 93.0 $\pm$ 3.8      & 96.0 $\pm$ 0.0        \\
SAM2Act (SAM2 $\rightarrow$ DINOv2) & 82.2 $\pm$ 0.5                   & 3.8                         & 97.0 $\pm$ 2.0          & 98.0 $\pm$ 2.3      & 69.0 $\pm$ 3.8      & \textbf{99.0} $\pm$ 2.0 & 80.0 $\pm$ 3.3       & 30.0 $\pm$ 7.7      & 89.0 $\pm$ 3.8      & 96.0 $\pm$ 0.0        \\
SAM2Act (Original Upsampling)       & 84.2 $\pm$ 0.9                   & 3.4                         &\textbf{100.0} $\pm$ 0.0 &\textbf{100.0} $\pm$ 0.0&\textbf{91.0} $\pm$ 3.8&99.0 $\pm$ 2.0      & 78.0 $\pm$ 9.5       & 29.0 $\pm$ 6.0      & 88.0 $\pm$ 5.7      & 96.0$\pm$ 0.0       \\
SAM2Act (w/o Multi-res Input)       & 85.7 $\pm$ 0.3                   & 2.7                         & 99.0 $\pm$ 2.0          & 96.0 $\pm$ 0.0      & 86.0 $\pm$ 8.3      & 98.0 $\pm$ 2.3          &\textbf{99.0} $\pm$ 2.0 & 43.0 $\pm$10.5      &\textbf{96.0} $\pm$ 0.0& \textbf{100.0} $\pm$ 0.0       \\
SAM2Act                             & \textbf{86.8} $\pm$ 0.5          & \textbf{2.3}                & 99.0 $\pm$ 2.0          & 99.0 $\pm$ 2.0      & 84.0 $\pm$ 5.7      & 98.0 $\pm$ 2.3          & 83.0 $\pm$ 6.0       &\textbf{47.0} $\pm$ 6.0& 93.0 $\pm$ 3.8      & \textbf{100.0} $\pm$ 0.0       \\
\toprule
\textbf{Method}                     & \textbf{Put in Cupboard} & \textbf{Put in Drawer} & \textbf{Put in Safe} & \textbf{Screw Bulb} & \textbf{Slide Block} & \textbf{Sort Shape} & \textbf{Stack Blocks} & \textbf{Stack Cups} & \textbf{Sweep to Dustpan} & \textbf{Turn Tap} \\
\midrule
SAM2Act (SAM2 $\rightarrow$ SAM)    & 72.0 $\pm$ 8.6           & 94.0 $\pm$ 2.3         &\textbf{99.0} $\pm$ 2.0 &\textbf{92.0} $\pm$ 5.7 & 97.0 $\pm$ 3.8       & 41.0 $\pm$ 3.8      & 73.0 $\pm$ 3.8        & 71.0 $\pm$ 2.0      & 96.0 $\pm$ 3.3            & 95.0 $\pm$ 2.0    \\
SAM2Act (SAM2 $\rightarrow$ Depth Anything V2) & 78.0 $\pm$ 2.3  & 96.0 $\pm$ 3.8         & 95.0 $\pm$ 2.0         & 90.0 $\pm$ 2.3         & 67.0 $\pm$ 2.0       & 45.0 $\pm$ 5.0      & 60.0 $\pm$ 3.3        & \textbf{91.0} $\pm$ 2.0   & \textbf{100.0} $\pm$ 0.0 & 90.0 $\pm$ 4.0    \\
SAM2Act (SAM2 $\rightarrow$ DINOv2) & \textbf{80.0} $\pm$ 3.3  & 99.0 $\pm$ 2.0         & 93.0 $\pm$ 3.8         & 89.0 $\pm$ 2.0         & 77.0 $\pm$ 2.0       & 49.0 $\pm$ 8.2      & 69.0 $\pm$ 5.0        & 79.0 $\pm$ 6.8      & 94.0 $\pm$ 2.3            & 92.0 $\pm$ 3.3    \\
SAM2Act (Original Upsampling)       & 69.0 $\pm$ 5.0           & 98.0 $\pm$ 2.3         & 96.0 $\pm$ 3.3        & 84.0 $\pm$ 3.3      &\textbf{99.0} $\pm$ 2.0   & 52.0 $\pm$ 3.3      & 71.0 $\pm$ 3.8        & 80.0 $\pm$ 3.3      & 99.0 $\pm$ 2.0            & 87.0 $\pm$ 6.0    \\
SAM2Act (w/o Multi-res Input)       & 72.0 $\pm$ 4.6           &\textbf{100.0} $\pm$ 0.0  & 96.0 $\pm$ 4.6      & 87.0 $\pm$ 2.0      & 82.0 $\pm$ 5.2       & 54.0 $\pm$ 5.2      & 74.0 $\pm$ 2.3            & 90.0 $\pm$ 6.9      & 97.0 $\pm$ 3.8            & 92.0 $\pm$ 4.6    \\
SAM2Act                             & 75.0 $\pm$ 3.8           & 99.0 $\pm$ 2.0         & 98.0 $\pm$ 2.3        & 89.0 $\pm$ 2.0      & 86.0 $\pm$ 4.0        &\textbf{64.0} $\pm$ 4.6&\textbf{76.0} $\pm$ 8.6  & 78.0 $\pm$ 4.0     & 99.0 $\pm$ 2.0            & \textbf{96.0} $\pm$ 5.7    \\

\bottomrule
\end{tabular}%
}
\label{appendix_tab:rlbench_abl}
\end{table*}

\subsection{\texttt{The Colosseum}}
\label{app:sam2act_ablation:colosseum}

We also conducted the same ablation experiments on \texttt{The Colosseum} generalization benchmark, as shown in \autoref{tab:colosseum}. The experimental setup remains the same as in \autoref{appendix_tab:rlbench_abl}, except that we did not test the variant of SAM2Act with the original convex upsampling. The results in \autoref{appendix_tab:rlbench_abl} show that removing SAM2’s multi-resolution image embedding inputs leads to a 14.8\% drop in performance, representing a relative decrease of 344.2\%. This highlights the effectiveness of SAM2’s multi-resolution image embeddings in providing robust visual representations, significantly enhancing SAM2Act’s generalization ability.

\section{RLBench Tasks}
\label{app:rlbench}

We follow the multi-task, multi-variation simulated experiment setup of PerAct \cite{shridhar2023perceiver}, RVT \cite{goyal2023rvt}, and RVT-2 \cite{goyal2024rvt}, using 18 RLBench tasks with 249 unique variations in object placement, color, size, category, count, and shape. A summary of the 18 RLBench tasks is provided in \autoref{appendix_tab:rlbench}. For a more detailed description of each task, please refer to PerAct \cite{shridhar2023perceiver}.

\begin{table*}[h!]
\centering
\caption{\textbf{The 18 RLBench Tasks for Multi-task Experiment.} We report on language template, the average number of extracted keyframes, the task variations, and the variation type.}
\resizebox{\textwidth}{!}{%
\begin{tabular}{@{}llccl@{}}

\toprule
Task name & Language Template & Avg. Keyframes & \#of Variations & Variation Type\\
\midrule     
put in drawer                   & “put the item in the \rule{0.3cm}{0.4pt} drawer”                                       &12.0  &3   &placement        \\
reach and drag                  & “use the stick to drag the cube onto the \rule{0.3cm}{0.4pt} target”                   &6.0   &20  &color        \\
turn tap                        & “turn \rule{0.3cm}{0.4pt} tap”                                                         &2.0   &2   &placement        \\
slide to target                 & “slide the block to \rule{0.3cm}{0.4pt} target”                                        &4.7   &4   &color        \\
open drawer                     & “open the \rule{0.3cm}{0.4pt} drawer”                                                  &3.0   &3   &placement        \\
put in cupboard                 & “put the \rule{0.3cm}{0.4pt} in the cupboard”                                          &5.0   &9   &category        \\
place in shape sorter           & “put the \rule{0.3cm}{0.4pt} in the shape sorter”                                      &5.0   &5   &shape        \\
put money in safe               & “put the money away in the safe on the \rule{0.3cm}{0.4pt} shelf”                      &5.0   &3   &placement        \\
push buttons                    & “push the \rule{0.3cm}{0.4pt} button, [then the \rule{0.3cm}{0.4pt} button]”           &3.8   &50  &color        \\
close jar                       & “close the \rule{0.3cm}{0.4pt} jar”                                                    &6.0   &20  &color         \\
stack block                     &  “stack \rule{0.3cm}{0.4pt} \rule{0.3cm}{0.4pt} blocks”                                &14.6  &60  &color,count  \\
place cups                      & “place \rule{0.3cm}{0.4pt} cups on the cup holder”                                     &11.5  &3   &count        \\
place wine at rack              & “stack the wine bottle to the \rule{0.3cm}{0.4pt} of the rack”                         &5.0   &3   &placement    \\
screw bulb                      & “screw in the \rule{0.3cm}{0.4pt} light bulb”                                          &7.0   &20  &color        \\
sweep to dustpan                &“sweep dirt to the \rule{0.3cm}{0.4pt} dustpan”                                         &4.6   &2   &size         \\
insert peg                      &“put the ring on the \rule{0.3cm}{0.4pt} spoke”                                         &5.0   &20  &color        \\
meat off grill                  &“take the \rule{0.3cm}{0.4pt} off the grill”                                            &5.0   &2   &category     \\
stack cups                      &“stack the other cups on top of the \rule{0.3cm}{0.4pt} cup”                            &10.0  &20  &color        \\

\bottomrule
\end{tabular}%
}
\label{appendix_tab:rlbench}
\end{table*}

\section{\texttt{MemoryBench} Tasks}
\label{app:memorybench}

In the following we provide details of the \texttt{MemoryBench} tasks.  

\subsection*{(a) Reopen drawer}

\textbf{Task Description:} 
The robot is instructed remember the drawer slot that was initially opened, and closed it and then press the button on the table, before finding back the previously opened drawer to re-open it.

\textbf{Success Metric:} 
The task is considered successful once the initial opened drawer has been re-opened.

\textbf{Objects:} 
A drawer and button.

\textbf{Variation Number:} 
3

\textbf{Keyframes:} 
8

\textbf{Language Instructions:}
\textit{"Close the drawer, then reopened the previously opened drawer while pushing the button in between."}

\bigskip

\subsection*{(b) Put block back}

\textbf{Task Description:} 
The robot is instructed move the block the centre, then push the button, then move the block back to its initial position.

\textbf{Success Metric:} 
The task is considered successful once the initial block has been moved back to its initial pose.

\textbf{Objects:} 
Four patch, one block and one button.

\textbf{Variation Number:} 
4

\textbf{Keyframes:} 
11

\textbf{Language Instructions:}
\textit{""Put the block to the centre and then back to its initial position while pushing the button in between.""}

\bigskip

\subsection*{(c) Rearrange block}

\textbf{Task Description:} 
The robot is instructed move the block in the centre to the empty patch, and then press the button, and then move the alternative block to the centre..

\textbf{Success Metric:} 
The task is considered successful once the alternative block has been moved to the centre.

\textbf{Objects:} 
Two patch, two blocks and one button.

\textbf{Variation Number:} 
2

\textbf{Keyframes:} 
10

\textbf{Language Instructions:}
\textit{"Move the block not on the patch to the empty patch, then press the button, then move the block that has not been moved off the patch."}

\section{\texttt{MemoryBench} Update}
\label{app:memorybench_update}

We updated the \texttt{reopen\_drawer} task in \texttt{MemoryBench} for the following reasons. During training on the original data, we observed that the gripper often collided with the drawer handle when closing the drawer. To prevent this, we introduced an additional waypoint for the closing motion, mirroring the procedure used for opening the drawer. Consequently, we retrained all policies specifically on this updated task. Furthermore, to standardize the memory window size across all three tasks, we also retrained SAM2Act+ on this task using a window size of 10, which led to improved performance. All results are updated to \autoref{tab:memorybench}.

\section{Real-world Experiments}
\label{app:real_world}

\begin{figure}
    \centering
    \includegraphics[width=\linewidth]{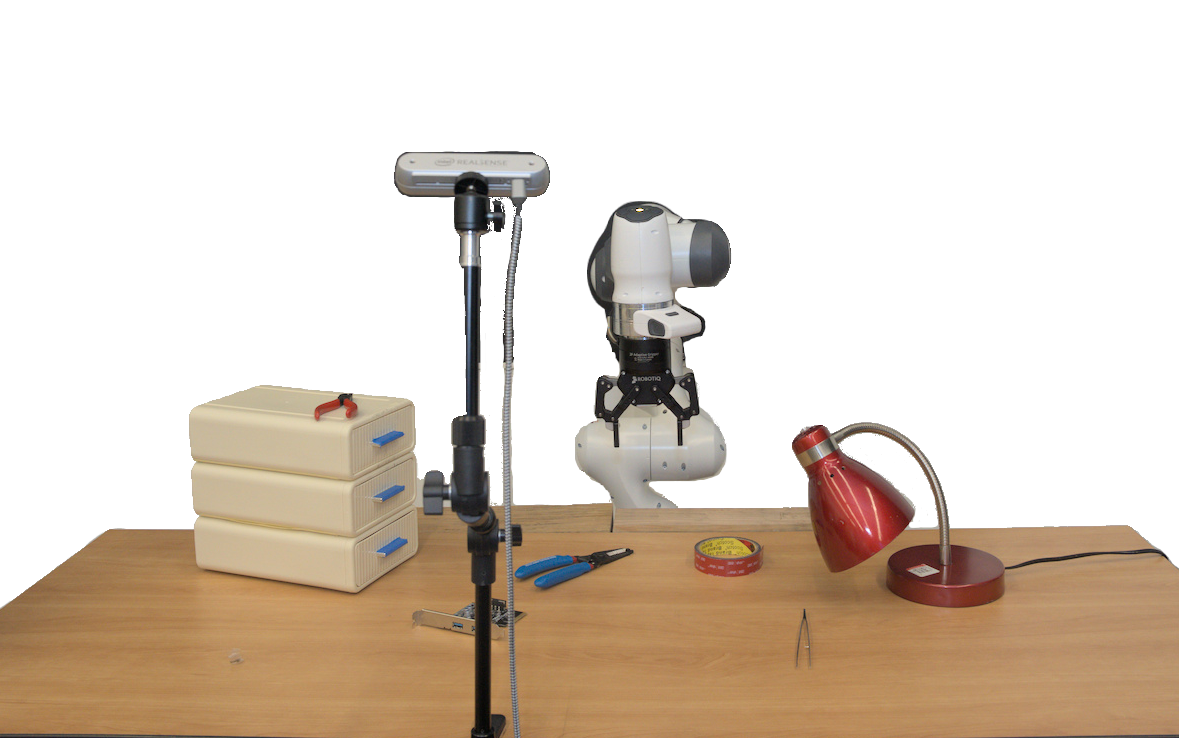}
    \caption{\textbf{Real-world Robot Setup.} A Franka Panda robot with a Robotiq Gripper. A RealSense D455 depth sensor captures the scene.}
    \label{fig:real_world_setup}
\end{figure}

In the following we provide details of the real-world setup and tasks. \autoref{fig:real_world_setup} illustrates the real-world setup. \autoref{tab:task_details_real_world} summarizes the properties of the real-world tasks.

\begin{table}[tp]
\centering
\caption{\textbf{Properties of the Real-world Tasks.} We report on language template, the average number of extracted keyframes, the number of items that the robot can interact with, the task variations, and the variation type.}
\label{tab:task_details_real_world}
\resizebox{\columnwidth}{!}{%
\begin{tabular}{@{}llcccc@{}}
\toprule
\multicolumn{1}{c}{Task name}                     & \multicolumn{1}{c}{Language template}    & \# keyframes & \# items & \# variations & variation type  \\ \midrule
(a) turn on the lamp             & “turn on the lamp”   & 4.5         & 1       & 1           & placement \\
(b) push buttons in sequence          & “push the red button, then the green button”     & 5         & 3       & 1      &     placement \\
(c) stack cubes      & “stack the \rule{0.3cm}{0.4pt} cube on the \rule{0.3cm}{0.4pt} cube”     & 4.0         & 5       & 3           & category,placement\\
(d) push the right button     & “push the button closest to the blue block”     & 6          & 3       & 1      & color,placement      \\

\bottomrule
\end{tabular}%
}

\end{table}

\subsection*{(a) Turn on the lamp}

\textbf{Task Description:} 
The robot is instructed to turn on a lamp by rotating its knob.

\textbf{Success Metric:} 
The task is considered successful once the lamp has been turned on by rotating the knob.

\textbf{Objects:} 
A single lamp.

\textbf{Coordination Challenges:} 
High precision is required to properly rotate the knob.

\textbf{Language Instructions:}
\textit{"Turn on the lamp."}

\bigskip

\subsection*{(b) Push buttons in sequence}

\textbf{Task Description:} 
The robot must press the red button first and then the blue button.

\textbf{Success Metric:} 
The task is considered successful if the buttons are pressed in the specified order: red, then blue. A third button is present but should remain unpressed.

\textbf{Objects:} 
Three buttons in front of the robot.

\textbf{Coordination Challenges:} 
Ensuring the robot presses the correct buttons in sequence without pressing the third button.

\textbf{Language Instructions:}
\textit{"Push the red button and then the blue button."}

\bigskip

\subsection*{(c) Stack blocks}

\textbf{Task Description:} 
The robot must place one specified block on top of another specified block.

\textbf{Success Metric:} 
The task is successful if the designated block is stacked on the correct target block.

\textbf{Objects:} 
Three single-colored blocks.

\textbf{Coordination Challenges:} 
Precision in picking and placing, plus correct language understanding to identify which block goes where.

\textbf{Language Instructions:}
\textit{"Stack the \textless item\textgreater{} block on the \textless item\textgreater{} block."}

\bigskip

\subsection*{(d) Push the same button}

\textbf{Task Description:} 
The robot must first identify and press the button closest to the blue block, then press the same button again after the block is removed.

\textbf{Success Metric:} 
The task is successful if the robot presses the correct button twice. Pressing the other button at any point results in failure.

\textbf{Objects:} 
Two buttons and one blue block (marking proximity).

\textbf{Coordination Challenges:} 
After the first button press, the blue block is removed; the robot must remember the button location to press it again.

\textbf{Language Instructions:}
\textit{"Push the button that is closest to the blue block. Press the same button again."}

\end{document}